%% file: neurips_2026.tex
\documentclass{article}

\PassOptionsToPackage{numbers, compress}{natbib}

\usepackage[main, final]{neurips_2026}

\usepackage[utf8]{inputenc} % allow utf-8 input
\usepackage[T1]{fontenc}    % use 8-bit T1 fonts
\usepackage{hyperref}       % hyperlinks
\usepackage{url}            % simple URL typesetting
\usepackage{booktabs}       % professional-quality tables
\usepackage{array}          % centered fixed-width table columns
\usepackage{longtable}      % multipage appendix tables
\usepackage{amsfonts}       % blackboard math symbols
\usepackage{nicefrac}       % compact symbols for 1/2, etc.
\usepackage{microtype}      % microtypography
\usepackage[table]{xcolor}  % colors
\usepackage{multirow}
\usepackage{graphicx}
\usepackage{amsmath}
\usepackage{pifont}
\usepackage{wrapfig}
\usepackage{enumitem}
\newcolumntype{C}[1]{>{\centering\arraybackslash}m{#1}}
\newcolumntype{L}[1]{>{\raggedright\arraybackslash}m{#1}}
\usepackage{makecell}

\usepackage{wasysym}

\DeclareRobustCommand{\cmpyes}{%
  \textcolor{green!50!black}{\ensuremath{\checkmark}}}
\DeclareRobustCommand{\cmpno}{%
  \textcolor{red!80!black}{\ensuremath{\times}}}
\DeclareRobustCommand{\cmppartial}{%
  \textcolor{orange!85!black}{\LEFTcircle}}
\title{AnyMo: Geometry-Aware Setup-Agnostic \\ Modeling of Human Motion in the Wild}

\author{%
  Baiyu Chen$^{1}$\thanks{Corresponding authors: breeze.chen@unsw.edu.au, flora.salim@unsw.edu.au.} \quad
  Zechen Li$^{1}$ \quad
  Wilson Wongso$^{1}$ \quad
  Lihuan Li$^{1}$ \quad
  Xiachong Lin$^{1}$ \\
  {\bfseries
  Hao Xue$^{1,2,3}$ \quad
  Benjamin Tag$^{1}$ \quad
  Flora Salim$^{1\ast}$} \\[4pt]
  $^1$The University of New South Wales \\
  $^2$The Hong Kong University of Science and Technology (Guangzhou) \\
  $^3$The Hong Kong University of Science and Technology
}

\begin{document}

\maketitle

\begin{abstract}
As wearable and mobile devices become increasingly embedded in daily life, they offer a practical way to continuously sense human motion in the wild. But inertial signals are highly dependent on the sensing setup, including body location, mounting position, sensor orientation, device hardware, and sampling protocol. This setup dependence makes it difficult to learn motion representations that transfer across devices and datasets, and limits the broader use of wearable IMUs beyond closed-set recognition. We introduce AnyMo, a geometry-aware framework for setup-agnostic human motion modeling. AnyMo uses physics-grounded IMU simulation over dense body-surface placements to generate diverse and plausible synthetic signals, pre-trains a graph encoder from paired synthetic placement views and masked partial observations, tokenizes multi-position IMU into full-body motion tokens, and aligns these tokens with an LLM for motion-language understanding. We evaluate AnyMo on three complementary tasks: zero-shot activity recognition across 14 unseen downstream datasets, cross-modal retrieval, and wearable IMU motion captioning, where it improves average Accuracy/F1/R@2 by 11.7\%/11.6\%/22.6\% on HAR, increases zero-shot IMU-to-text and text-to-IMU retrieval MRR by 15.9\% and 28.6\%, respectively, and improves zero-shot captioning BERT-F1 by 7.5\%. These results support AnyMo as a generalist model for wearable motion understanding in the wild. Project page: \url{https://anymo-project.github.io}.
\end{abstract}

\input{sec/1_introduction}
% \input{sec/2_related_works}
\input{sec/3_method}
\input{sec/4_experiments}
\input{sec/5_conclusion}

\begin{ack}
This research includes computations using the Wolfpack computational cluster, supported by the School of Computer Science and Engineering at UNSW Sydney. We also acknowledge support from the ARC Centre of Excellence for Automated Decision-Making and Society (CE200100005). Hao Xue would like to thank HKUST-GZ Startup Fund G0101000338 for its support.
\end{ack}

% \section*{References}

\bibliographystyle{plainnat}
\bibliography{ref}

% References follow the acknowledgments in the camera-ready paper. Use unnumbered first-level heading for
% the references. Any choice of citation style is acceptable as long as you are
% consistent. It is permissible to reduce the font size to \verb+small+ (9 point)
% when listing the references.
% Note that the Reference section does not count towards the page limit.
% \medskip

% {
% \small

% [1] Alexander, J.A.\ \& Mozer, M.C.\ (1995) Template-based algorithms for
% connectionist rule extraction. In G.\ Tesauro, D.S.\ Touretzky and T.K.\ Leen
% (eds.), {\it Advances in Neural Information Processing Systems 7},
% pp.\ 609--616. Cambridge, MA: MIT Press.

% [2] Bower, J.M.\ \& Beeman, D.\ (1995) {\it The Book of GENESIS: Exploring
%   Realistic Neural Models with the GEneral NEural SImulation System.}  New York:
% TELOS/Springer--Verlag.

% [3] Hasselmo, M.E., Schnell, E.\ \& Barkai, E.\ (1995) Dynamics of learning and
% recall at excitatory recurrent synapses and cholinergic modulation in rat
% hippocampal region CA3. {\it Journal of Neuroscience} {\bf 15}(7):5249-5262.
% }

%%%%%%%%%%%%%%%%%%%%%%%%%%%%%%%%%%%%%%%%%%%%%%%%%%%%%%%%%%%%

\newpage

\appendix

\section{Experiments and Implementation Details}
\label{sec:experiment_details}
\subsection{Training Data and Window Construction}
\label{sec:training_data_window_construction}
All AnyMo training stages use Nymeria~\citep{ma2024nymeria} as the only source of motion-text supervision.
We first synchronize Nymeria body motion, mesh vertices, atomic-action annotations, and real IMU streams to a common 60 Hz timeline.
Most exported instruction-tuning rows correspond to 5 s windows: after removing IMU boundary tokens, 90.1\% of rows contain 150 IMU code tokens, which corresponds to $T=300$ frames because the original frame count is twice the IMU code-token count.
The remaining text-aligned rows keep their native atomic-action durations, so $T$ can be shorter or longer.
Each frame contains six IMU channels (three-axis acceleration and three-axis angular velocity), and each window is represented on a fixed 23-node body graph following the Xsens kinematic tree: pelvis, spine, neck/head, both shoulders/arms/hands, and both legs/feet/toes.
The graph representation has shape $6 \times T \times 23 \times 1$.
For sparse wearable observations, only the nodes corresponding to the visible sensors are exposed to the model; the other graph nodes are replaced by the learned mask token in the encoder.

Before applying the held-out-subject protocol, the text-aligned Nymeria export contains 828 recording samples and 168{,}295 windows.
We choose the held-out subjects so that the held-out split remains subject-disjoint while still covering all 20 original Nymeria scenarios.
Specifically, we reserve five subjects (\texttt{alec\_meza}, \texttt{bradley\_herman}, \texttt{dominique\_frye}, \texttt{justin\_ramirez}, and \texttt{kyle\_parker}), whose held-out recordings collectively cover the 20 scenarios, and exclude these recordings from the text-aligned token and instruction-tuning exports.
The resulting subject-disjoint IMU-token pre-training corpus contains 808 Nymeria recording samples and 164{,}387 text-aligned windows.
The instruction-tuning export contains 159{,}098 labeled windows and expands them into 986{,}322 language-model instruction rows, evenly split between narration and activity multiple-choice tasks.
The same subject-disjoint export provides 986{,}322 paired IMU-text contrastive rows, using the original atomic-action narration and conservative augmented narrations as positive texts.

\subsection{Synthetic IMU Generation}
\label{sec:synthetic_imu_generation_details}
The synthetic pre-training signals are generated from Nymeria's anatomically grounded human model, using its mesh and skeleton motion.
For every body segment, we build a surface template from selected mesh vertices.
At each candidate vertex, we estimate a local right-handed sensor frame from the surface normal, an anatomical tangent direction, and the corresponding binormal.
The local offset between the posed mesh vertex and the segment coordinate frame is then tracked through the motion sequence, producing a dense set of geometry-aware candidate IMU placements.
We define our surface selection from the template-mesh skinning weights.
For each mesh vertex, we sort the influencing skeleton joints by skinning weight and assign the vertex to a body segment if one of that segment's joints appears among the top two nonzero influences.
This is a compromise between two less useful extremes: a strict top-1 assignment can make shared boundary vertices disappear from nearby segments, while treating every positive skinning weight as a candidate can spread each segment over overly broad and weakly related surface regions.
The resulting selected vertex counts are shown in \autoref{tab:selected_surface_vertices}; in total, the exported synthetic archive contains 831 eligible Nymeria samples, each with 2{,}374 candidate placements across the 23 body segments.

\begin{table}[h]
\caption{Selected surface vertices per body segment for synthetic IMU placement.}
\label{tab:selected_surface_vertices}
\centering
\begin{tabular}{lrlrlr}
\toprule
Segment & \# Vertices & Segment & \# Vertices & Segment & \# Vertices \\
\midrule
Pelvis & 108 & T8 & 105 & Right Hand & 145 \\
L5 & 59 & Neck & 44 & Left Shoulder & 45 \\
L3 & 35 & Head & 290 & Left Upper Arm & 90 \\
T12 & 90 & Right Shoulder & 42 & Left Forearm & 120 \\
Right Upper Arm & 90 & Left Hand & 145 & Left Upper Leg & 106 \\
Right Forearm & 127 & Right Upper Leg & 108 & Left Lower Leg & 122 \\
Right Lower Leg & 122 & Right Foot & 125 & Left Foot & 112 \\
Right Toe & 72 & Left Toe & 72 &  &  \\
\bottomrule
\end{tabular}
\end{table}

During representation pre-training, each training window is sampled twice to form two full-body graph views.
For each segment in each view, one candidate placement is selected independently.
We further apply surface rotation augmentation, with an in-plane rotation range of $\pm 180^\circ$ and a small tilt range of $\pm 10^\circ$ during training.
Acceleration and angular velocity are computed from the simulated rigid-body trajectory in the candidate sensor frame.
Real Nymeria IMU streams from the head and wrists are not used as training targets; they are used to estimate device-noise priors and to define the held-out sim-to-real evaluation protocol.
We estimate this noise prior from the two Nymeria device streams and three real wearable sites to make the synthetic IMU signals more realistic.

\begin{figure}[h]
    \centering
    \includegraphics[width=\linewidth]{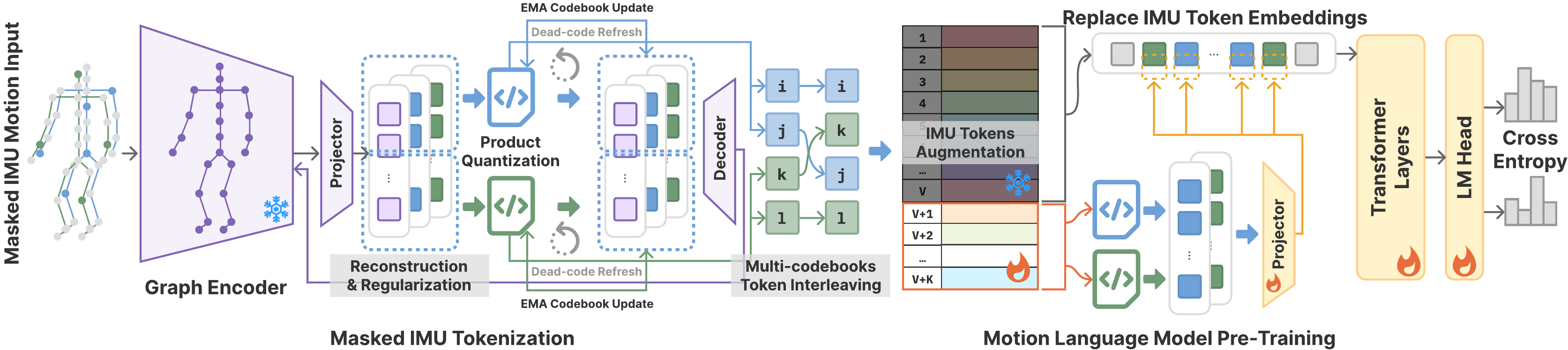}
    \caption{More details of Masked IMU Tokenization and Motion Language Model Pre-Training.}
    \label{fig:tokenization_pretraining_detail}
\end{figure}

\subsection{Representation Pre-Training and Tokenization}
\label{sec:representation_pretraining_details}
As illustrated in \autoref{fig:tokenization_pretraining_detail}, the IMU encoder is an ST-GCN~\citep{yan2018spatial} over the 23-node body graph.
It uses ten spatio-temporal graph convolution blocks with temporal kernel size 9, channel widths increasing from 64 to 128 to 256, and two temporal stride-2 stages.
Thus a 300-frame input window is encoded as a time-preserving latent sequence of length about 75, with latent dimension 256.
Pre-training samples two independent full-body graph views from the synthetic candidate set and two corresponding sparse views.
For each sparse view, a random number of visible nodes between 1 and 5 is retained, and all remaining nodes are masked.
A six-layer Transformer~\citep{vaswani2017attention} predictor maps the sparse-view latent sequence to the opposite full-view target sequence.
We train with the symmetric predictive InfoNCE objective, temperature 0.1, and stop-gradient targets.
Unless otherwise noted, the ST-GCN optimizer is AdamW with learning rate $3\times10^{-4}$, batch size 64, and 10 epochs.
The ST-GCN encoder pre-training is run on a single NVIDIA L40S GPU.

\begin{table}[h]
\caption{Final PQ-VAE tokenizer diagnostics for the exported IMU tokenizer. Perplexity and top-$k$ code mass summarize codebook usage, while exact sequence collision is measured after interleaving the two codebook streams.}
\label{tab:pqvae_tokenizer_diagnostics}
\centering
\begin{tabular}{lcc}
\toprule
Metric & Codebook 0 & Codebook 1 \\
\midrule
Used codes & 2{,}033 / 2{,}048 & 2{,}046 / 2{,}048 \\
Usage rate & 99.27\% & 99.90\% \\
Dead-code ratio & 0.73\% & 0.10\% \\
Perplexity & 1{,}285.7 & 1{,}098.1 \\
Top-1 code mass & 1.46\% & 1.11\% \\
Top-10 code mass & 5.17\% & 4.96\% \\
\midrule
Exact sequence collision rate & \multicolumn{2}{c}{0.61\%} \\
\bottomrule
\end{tabular}
\end{table}

After ST-GCN pre-training, we freeze the encoder and train a product-quantized VAE on masked sparse-view latent sequences.
The tokenizer uses two codebooks with 2{,}048 entries each, a 128-dimensional bottleneck, and 64-dimensional code vectors per codebook.
The product-quantizer codebooks are updated with exponential-moving-average (EMA) statistics using decay 0.99.
To prevent unused IMU tokens from persisting, we apply dead-code refresh after each EMA update: within each codebook, entries whose accumulated usage falls below 20\% of the average code usage are replaced by latent vectors sampled from the current mini-batch.
The decoder reconstructs the frozen ST-GCN latent sequence with a SmoothL1 reconstruction loss plus the standard commitment loss.
At export time, the two codebook indices are interleaved over time.
Consequently, a 5 s Nymeria window normally yields 75 latent steps and 150 IMU code tokens, plus boundary tokens.
We add 4{,}096 IMU code tokens and IMU boundary tokens to the LLM tokenizer, while keeping the learned codebook vectors as the continuous representation behind those discrete token IDs.
We report final tokenizer diagnostics in \autoref{tab:pqvae_tokenizer_diagnostics}, including codebook usage, concentration statistics, and exact sequence collisions.
These diagnostics check the tokenizer at two levels.
At the codebook level, the near-complete usage rates and low dead-code ratios show that the product quantizer does not collapse to a small subset of entries.
The perplexities, 1{,}285.7 and 1{,}098.1 out of 2{,}048 entries, indicate broad but non-uniform code usage, while the low top-1 and top-10 masses show that no small group of codes dominates the assignment distribution.
At the sequence level, the exact sequence collision rate is computed after interleaving the two codebook streams over time; the 0.61\% collision rate indicates that almost all tokenized IMU windows receive distinct discrete token sequences.

\subsection{Motion-Language Training}
\label{sec:motion_language_training_details}
AnyMo uses Qwen2.5-0.5B~\citep{qwen2025qwen25technicalreport} as the language backbone.
We implement the motion-language pre-training stage with ms-swift~\citep{zhao2025swift}, using a custom AnyMo model registration that replaces IMU token embeddings with projected codebook vectors at runtime.
All motion-language training stages are run on two NVIDIA L40S GPUs.
The LLM is first adapted with an IMU-token language-modeling stage in which the assistant target consists of IMU token sequences exported from the frozen ST-GCN and tokenizer.
The IMU code-token embeddings are not ordinary lookup rows: at runtime, each IMU token ID is mapped back to its product-quantizer code vector and projected into the Qwen hidden space by a two-layer MLP.
The corresponding language-model head rows are initialized from the same projected code vectors.
For this stage, we use full fine-tuning with \texttt{bf16}, a maximum sequence length of 1024, learning rate $10^{-4}$, batch size 16, and 3 epochs.

Before instruction tuning, we augment the Nymeria atomic-action narrations with GPT-OSS-120B~\citep{openai2025gptoss120bgptoss20b}.
For each atomic-action text, we treat the original annotation as the ground-truth semantic anchor and generate five paraphrases that preserve the same action meaning while varying wording, conciseness, temporal emphasis, body-motion emphasis, and scene generality.
We then run a GPT-OSS-120B self-verifier that checks whether the five variants preserve the atomic action, maintain the action order, avoid unobserved intent or invented content, and provide sufficient diversity.
If the verifier rejects the whole set, we regenerate all five variants; if fewer than five variants fail, we repair only the rejected variants while keeping the accepted ones.
The verifier is applied after each generation or repair step, with at most two additional regeneration/repair rounds, giving at most three verifier passes per atomic-action row.

For the label-contrastive branch, we construct a closed activity-label vocabulary from the same atomic-action annotations.
We first ask GPT-OSS-120B to assign a free-form activity label to every atomic action.
Because these free-form labels can differ substantially in surface form, we normalize them by lowercasing, removing punctuation and extra spaces, and canonicalizing leading articles such as ``a'', ``an'', and ``the'' before exact deduplication.
We embed labels with Qwen3-Embedding-8B~\citep{zhang2025qwen3textembedding} and cluster labels with frequency at least five using a cosine-similarity threshold of 0.85.
For clusters containing two or more labels, GPT-OSS-120B merges semantically close labels into either one simple label or a slash-separated label when multiple equivalent phrasings should be retained.
Human experts review roughly 1{,}000 candidate labels, remove ambiguous categories, and finalize AnyMo-180, a 180-class activity-label vocabulary.
The resulting AnyMo-180 label corpus turns Nymeria into one of the largest fine-grained IMU-based HAR training corpora, providing an activity vocabulary that goes beyond the small closed label spaces common in wearable HAR and helps mitigate coarse activity labels and fragmented activity vocabularies~\citep{haresamudram2025past, cai2025towards, bian2026foundation}.
To label windows against AnyMo-180, we run GPT-5.4 nano twice as an enum-text classifier over the 180 classes.
Rows with identical labels across the two runs are accepted directly.
Disagreements are adjudicated by GPT-5.4 mini, and we retain only labels with a majority vote after adjudication.
The two GPT-5.4 nano passes agree on 128{,}461 of 168{,}295 rows (76.33\%); among the 39{,}834 disagreements, GPT-5.4 mini produces a majority label for 34{,}873 rows.
After this filtering, 163{,}334 rows remain labeled, corresponding to 97.05\% of the 168{,}295 candidate rows.
We then perform a second class-wise audit over the curated label set to further improve label quality.
For each candidate row, we use GPT-5.5 High to compare the assigned activity label against the current source-text annotation and the preceding and following source-text annotations to verify the local temporal context; inconsistent rows are relabeled or dropped under human-expert supervision.
This final audit yields 158{,}138 labeled rows, corresponding to 93.96\% of candidate rows. The class distribution is long-tailed, reflecting in-the-wild activity frequencies: per-class counts range from 18 to 19{,}696 rows, with a median of 342.5 and a mean of 878.54.
The final label counts are reported in \autoref{tab:final_activity_label_counts}.

\begin{table}[p]
\caption{AnyMo-180 activity-label vocabulary and counts.}
\label{tab:final_activity_label_counts}
\centering
\tiny
\setlength{\tabcolsep}{3pt}
\begin{tabular}{@{}p{0.18\textwidth}r p{0.18\textwidth}r p{0.18\textwidth}r p{0.18\textwidth}r@{}}
\toprule
Label & Count & Label & Count & Label & Count & Label & Count \\
\midrule
walking / hiking & 19{,}696 & standing & 7{,}411 & sitting & 5{,}906 & play board games / cards / puzzle & 5{,}873 \\
placing items on a surface & 4{,}758 & picking up objects & 4{,}248 & slicing / chopping / cutting / dicing food & 3{,}765 & folding clothes / pants / towel / blanket & 3{,}752 \\
gesturing & 3{,}716 & dancing & 3{,}254 & kneeling & 3{,}130 & turning in place / turning / spinning & 2{,}584 \\
playing soccer / kicking ball & 2{,}522 & closing refrigerator / cabinet / drawer / closet / door & 2{,}493 & playing charades & 2{,}476 & playing badminton & 2{,}419 \\
eating & 2{,}323 & standing up and walking & 2{,}239 & walking while carrying objects & 2{,}141 & opening and closing refrigerator / cabinet / drawer / closet / door & 2{,}116 \\
using laptop & 2{,}058 & raising hand(s) / arm(s) & 1{,}987 & adjusting pillows / sofa pillows & 1{,}979 & pulling a drawer & 1{,}941 \\
side stepping exercise & 1{,}887 & opening refrigerator / cabinet / drawer / closet / door & 1{,}828 & yoga / stretching & 1{,}778 & placing items into a box / bag / container & 1{,}584 \\
watching tv & 1{,}581 & arranging decorations & 1{,}536 & closing refrigerator / cabinet / drawer / closet / door and walking & 1{,}470 & playing air hockey / playing table hockey / playing foosball & 1{,}433 \\
walking down stairs & 1{,}345 & walking up stairs & 1{,}311 & squatting & 1{,}253 & biking outdoors & 1{,}165 \\
writing & 1{,}146 & opening refrigerator / cabinet / drawer / closet / door and walking & 1{,}114 & throwing objects & 1{,}036 & lying & 976 \\
scooping food & 971 & making coffee & 910 & playing arcade game & 882 & drinking & 872 \\
drawing & 857 & marching in place & 822 & measuring appliance / furniture / door & 802 & stirring / mixing & 795 \\
hanging clothing & 780 & disposing trash & 779 & moving chair & 703 & spreading / unfolding blanket / clothes & 664 \\
twisting torso & 635 & scratching / touching face & 623 & adding food to container / pan & 608 & loading dishwasher / unloading dishwasher & 601 \\
adjusting blinds / adjusting curtains & 593 & rotating hands / arms & 586 & opening bag / package / pack / pouch & 576 & inspecting appliance / furniture & 569 \\
walking / stepping in place & 555 & cleaning tableware / kitchenware & 549 & stir-frying & 544 & lunges / lunging & 541 \\
playing table tennis & 539 & using phone & 538 & standing up & 498 & opening refrigerator / cabinet / drawer / closet and taking objects & 477 \\
adjusting stove / turning on stove / opening stove / turning stove knob & 457 & knee raises & 447 & sweeping & 445 & spreading sauce / cream / butter on bread / food & 444 \\
turning off / on light switch & 441 & turning on faucet & 441 & playing billiards & 425 & walking while picking up object & 418 \\
adjusting glasses & 414 & tiptoeing & 412 & opening a box / container & 410 & cleaning whiteboard & 404 \\
sitting down & 398 & poking food & 393 & folding / rolling / tearing / peeling tape & 381 & punching / shadowboxing & 377 \\
reading & 374 & seasoning food & 367 & walking with bike & 353 & running & 352 \\
turning / shaking / nodding head & 346 & looking under furniture & 345 & closing a box / container & 340 & jumping / hopping in place & 319 \\
looking around & 319 & playing catch & 314 & butt kicks & 312 & operating electric kettle & 306 \\
playing darts & 302 & stacking items & 283 & flipping food in pan & 279 & capping marker / bottle / jar & 268 \\
stationary biking / riding an exercise bike & 267 & opening bottle / jar & 264 & vacuuming / walking while vacuuming & 256 & washing hands & 250 \\
moving sofa / couch & 249 & cleaning a surface & 243 & jogging in place & 243 & standing foot rotation / ankle rotation & 243 \\
operating oven & 240 & pressing / mashing food in pan & 240 & pouring liquid & 234 & using microwave & 224 \\
peeling fruits / vegetable & 223 & moving table & 221 & alternating punches while stepping in place & 218 & foot tapping & 217 \\
inserting pillow into pillowcase & 216 & placing pan on stove & 216 & swinging on a swing & 215 & pulling blinds / pulling curtains & 208 \\
filling a container with liquid & 195 & organizing books / shelves & 193 & marching & 187 & front kicks & 182 \\
air writing & 169 & standing leg raise & 168 & drying hands & 161 & using washing machine & 154 \\
jogging & 142 & playing with rubik's cube & 140 & eating while playing board game & 137 & pouring food into container & 137 \\
plank / alternating plank & 130 & whisking / beating eggs & 126 & cracking egg & 116 & bicep curls & 115 \\
organizing kitchen items & 114 & stepper exercise / using stepper & 114 & moving bed & 111 & ab rollout exercise & 110 \\
organizing clothes / towels & 110 & untangling party decorations & 110 & kneading dough / shaping dough & 109 & adjusting clothing & 108 \\
making sandwich & 89 & seated leg raises & 89 & adjusting hair & 85 & catching objects & 83 \\
running in place & 83 & rolling a yoga ball & 82 & adjusting headset & 80 & playing with a pool noodle & 80 \\
mopping & 77 & using toaster & 76 & bowling & 72 & walking while drinking & 72 \\
checking smartwatch & 69 & opening trash bin & 69 & clapping & 68 & jumping rope & 68 \\
tearing paper towel / tearing tissue & 68 & pulling chair and sitting & 65 & wiping mouth with tissue / napkin / paper towel & 61 & removing seeds from food & 59 \\
tying shoe laces & 57 & walking while punching & 51 & playing toy & 50 & standing arm circles & 47 \\
washing fruit / vegetables & 46 & writing with foot & 45 & dumbbell lateral raises & 43 & setting up pool table & 42 \\
hanging decorations / frames & 39 & threading beads & 38 & standing elbow-to-knee crunch & 35 & sit-ups & 34 \\
sitting up & 34 & playing rock-paper-scissors & 33 & pruning flower & 21 & squeezing fruit & 18 \\
\bottomrule
\end{tabular}
\end{table}

We then perform instruction tuning with both language-modeling and contrastive losses.
The language-modeling branch contains two task families: IMU-to-narration generation and IMU-to-activity multiple choice.
For each multiple-choice example, the correct activity is mixed with a fixed 35-choice candidate set sampled from the training activity pool.
The contrastive branch encodes IMU prompts and text prompts with the same Qwen backbone, then applies branch-specific latent-attention poolers and projection heads before a symmetric IMU-text contrastive loss.
The reported contrastive checkpoint uses an 8-token learnable soft prompt on the text/label branch.
We train the contrastive instruction-tuning stage for 1 epoch with maximum sequence length 4096, learning rate $2\times10^{-5}$, contrastive temperature 0.1, language-modeling loss weight 1.0, and contrastive loss weight 2.0.
For contrastive batches, we use a per-GPU batch size of 16 and 4 GradCache steps~\citep{gao2021scaling}; with NVIDIA L40S GPUs, this gives an effective global contrastive batch size of $16 \times 4 \times 2 = 128$.

\subsection{Zero-Shot Recognition Evaluation}
\label{sec:zero_shot_recognition_eval_details}
Downstream HAR datasets are used only for evaluation.
All dataset adapters convert accelerometer channels to $\mathrm{m/s^2}$ and gyroscope channels to $\mathrm{rad/s}$ when the source units are known, resample windows to 60 Hz, and map each physical device location to the closest node(s) in the 23-node body graph. This mapping requires only the body segment on which each device is worn, information typically known in everyday use or provided in dataset metadata,
without requiring precise sensor-coordinate measurements or calibration. For head-mounted video datasets such as Ego4D, EgoExo4D, and MMEA, the visible node is \texttt{Head}.
For wrist, arm, waist, torso, leg, and foot sensors, the adapters expose the corresponding upper-limb, spine, or lower-limb graph nodes.

Each evaluation example is tokenized by the frozen ST-GCN and tokenizer, then scored without updating AnyMo.
For recognition, we encode the IMU prompt and each candidate activity label prompt, compute cosine similarities in the learned contrastive space, and rank all labels from the dataset's label vocabulary.
Accuracy is computed from the top-ranked label, macro-F1 is computed over predicted labels, and Recall@2 is positive if the ground-truth label appears in the top two ranked labels.
The label vocabulary is dataset-specific and includes all activities for that benchmark split, so the evaluation does not require dataset-specific classifier heads.

\subsection{Retrieval and Captioning Evaluation}
\label{sec:retrieval_captioning_eval_details}
For sim-to-real evaluation on Nymeria, we export held-out examples from the five reserved subjects listed above.
This produces 20 held-out recording samples and 3{,}908 text-aligned windows for both retrieval and captioning.
For OOD evaluation on EgoExo4D, we construct a balanced 4{,}000-window atomic-action subset with 500 examples from each of eight parent tasks: rock climbing, basketball, dance, cooking, health, bike repair, soccer, and music.
These windows are sampled from EgoExo4D's 200 Hz head-IMU streams, resampled to 60 Hz, and tokenized with the same frozen AnyMo tokenizer.

Retrieval uses the same prompt templates as contrastive instruction tuning.
For IMU-to-text retrieval, each IMU embedding is compared against the pool of primary reference narrations; for text-to-IMU retrieval, each text embedding is compared against the pool of IMU embeddings.
We report Recall@1, Recall@5, Recall@10, and MRR in both directions.
Captioning uses the IMU-to-narration prompt and greedy decoding with a maximum of 64 generated tokens.
Generated captions are evaluated against the primary atomic-action narration using BLEU, ROUGE-L, METEOR, and BERT-F1.
For BERT-F1, we compute it with the \texttt{deberta-xlarge-mnli}~\citep{he2021deberta}~\footnote{\url{https://huggingface.co/microsoft/deberta-xlarge-mnli}}.

\subsection{Capability Radar Details}
\label{sec:capability_radar_details}
\autoref{fig:method_families} summarizes the main results with nine radar axes drawn from the three evaluation groups.
Exact Activity Match, Balanced Recognition, and Top-2 Activity Recall correspond to the average zero-shot HAR Accuracy, macro-F1, and Recall@2 across the 14 downstream recognition datasets.
IMU-to-Text Rank Quality and Text-to-IMU Rank Quality correspond to MRR for the two retrieval directions on the EgoExo4D zero-shot 100-sample candidate-ranking split.
Word Match, Sequence Overlap, Content Alignment, and Semantic Similarity correspond to BLEU-1, ROUGE-L, METEOR, and BERTScore-F1 on the EgoExo4D zero-shot captioning split.
Because these metrics have different natural ranges, the radar chart rescales each axis independently for visualization; the axis tick labels report the original metric values.

\subsection{Ablation Implementation Details}
\label{sec:ablation_implementation_details}
All ablations in \autoref{tab:ablation} use the same downstream zero-shot recognition protocol, candidate label vocabularies, and metrics as the full AnyMo model.
When an ablation changes an upstream stage, all downstream artifacts that depend on that stage are regenerated, including the tokenizer export, IMU-token pre-training rows, instruction-tuning rows, and evaluation tokens.
This keeps the comparison focused on the removed component rather than mixing incompatible tokenizers or encoders.

\textbf{Geometry-aware simulation.}
The full model uses geometry-aware synthetic IMU generated from mesh-surface candidate placements, with local sensor frames estimated from surface normals and anatomical tangents as described in \autoref{sec:synthetic_imu_generation_details}.
For \textit{w/o geometry-aware simulation}, we instead generate synthetic IMU from a joint-mounted setup commonly used in prior synthetic-IMU work: one default virtual sensor is rigidly attached to each of the 23 Xsens segment frames.
This removes mesh-vertex candidate placements, local surface frames, normal/tangent-based orientation variation, surface-rotation augmentation, and real-device calibration.
All later stages are rerun from this synthetic data: the ST-GCN encoder is trained with the same masked cross-view predictive InfoNCE objective, the PQ-VAE tokenizer is retrained on the resulting sparse-view latents, the motion-language pre-training and instruction-tuning data are re-exported with that tokenizer, and the final AnyMo model is evaluated with the same recognition protocol.
Because the joint-mounted setup provides only one placement per body node, it removes within-node device-placement diversity; this ablation tests whether conventional joint-level synthetic IMU is sufficient without dense geometry-aware setup variation.

\textbf{Masked cross-view predictive contrastive pre-training.}
The full encoder objective samples two full-body graph views, samples sparse visible-node masks for each view, and trains a Transformer predictor to map each sparse-view latent sequence to the opposite full-view latent sequence with symmetric InfoNCE and stop-gradient targets.
For \textit{w/o masked cross-view pred. contrastive}, we keep the same synthetic graph-view sampling but replace this sparse-to-full predictive objective with a standard full-view contrastive objective.
Concretely, the two independently sampled full graph views are encoded, projected, and contrasted directly against each other with sequence-level InfoNCE.
No sparse visible-node mask and no sparse-to-full predictor are used in this encoder-pretraining ablation.
The tokenizer, motion-language pre-training rows, instruction-tuning rows, and downstream evaluation tokens are then regenerated from this encoder.

\textbf{Motion-language contrastive losses.}
The full contrastive instruction-tuning stage combines three losses: the language-modeling loss on instruction rows, a narration-level symmetric IMU-text contrastive loss between IMU prompts and atomic-action narration prompts, and a supervised label-level contrastive loss between IMU prompts and curated activity-label prompts.
For \textit{w/o label contrastive}, we set the label-level contrastive loss weight to zero while keeping the narration-level contrastive loss and the language-modeling loss unchanged.
For \textit{w/o narration contrastive}, we set the narration-level contrastive loss weight to zero while keeping the label-level contrastive loss and the language-modeling loss unchanged.
For \textit{w/o all contrastive}, we remove both contrastive branches and train only with the language-modeling instruction-tuning objective from the IMU-token pre-trained checkpoint.

\textbf{MCQ instruction tuning.}
The full language-modeling branch is balanced between IMU-to-narration rows and IMU-to-activity multiple-choice rows.
For \textit{w/o MCQ instruction tuning}, we remove all MCQ instruction rows and replace them with additional narration rows, so that the total number of language-modeling instruction examples remains unchanged.
The contrastive instruction-tuning losses and all other hyperparameters are unchanged for this ablation.

\subsection{Setup Transfer and Real-Synthetic Alignment}
\label{sec:setup_transfer_alignment}

\subsubsection{Cross-Dataset Evaluation on CIP-MTwAwinda}

\textbf{Evaluation protocol.}
We further evaluate real-synthetic correspondence on CIP-MTwAwinda~\citep{palermo2022cipdatabase}, which provides synchronized motion and real accelerometer and gyroscope measurements at 17 body locations.
We use the fixed AnyMo model without adaptation to this dataset and construct 1,000 paired motion windows.
For each window and observed segment, we sample a corresponding synthetic body-surface placement.
Seven wearable configurations cover the head, torso, upper limbs, pelvis, and lower limbs.
We report mean$\pm$standard deviation over 20 sampling seeds.
Bidirectional exact-pair retrieval uses all 1,000 opposite-domain windows as candidates.

\begin{table}[h]
\caption{Zero-shot paired real-synthetic alignment on CIP-MTwAwinda across 17 real sensor locations and seven wearable setups. Results are mean{\tiny$\pm$std} over 20 sampling seeds. Paired retrieval uses all 1,000 synchronized windows as candidates and is averaged over both directions.}
\label{tab:cip_setup_alignment}
\centering
\scriptsize
\setlength{\tabcolsep}{1.2pt}
\renewcommand{\arraystretch}{1.05}
\resizebox{\linewidth}{!}{%
\begin{tabular}{@{}l|*{6}{c}|*{3}{c}@{}}
\toprule
Real Sensor Setup
& \makecell[c]{Cos. Sim.$\,\uparrow$}
& \makecell[c]{Domain Classifier\\Acc. (\%)$\,\to 50$}
& \makecell[c]{kNN Domain\\Mixing (\%)$\,\to 50$}
& \makecell[c]{MMD$^2\,\downarrow$}
& \makecell[c]{Domain\\Silhouette$\,\to 0$}
& \makecell[c]{Domain\\Purity (\%)$\,\to 50$}
& \makecell[c]{Paired\\R@1 (\%)$\,\uparrow$}
& \makecell[c]{Paired\\R@5 (\%)$\,\uparrow$}
& \makecell[c]{Paired\\MRR (\%)$\,\uparrow$} \\
\midrule
Head + Hands
& 0.832{\tiny$\pm$0.001}
& 63.5{\tiny$\pm$1.0}
& 47.2{\tiny$\pm$0.3}
& 0.0038{\tiny$\pm$0.0002}
& 0.005{\tiny$\pm$0.002}
& 51.2{\tiny$\pm$0.3}
& 48.9{\tiny$\pm$0.9}
& 71.2{\tiny$\pm$0.6}
& 59.2{\tiny$\pm$0.6} \\
\midrule
Sternum + Pelvis
& 0.775{\tiny$\pm$0.001}
& 87.8{\tiny$\pm$0.6}
& 35.8{\tiny$\pm$0.3}
& 0.0096{\tiny$\pm$0.0001}
& 0.013{\tiny$\pm$0.001}
& 52.3{\tiny$\pm$0.7}
& 37.3{\tiny$\pm$0.7}
& 64.4{\tiny$\pm$0.7}
& 49.9{\tiny$\pm$0.5} \\
\midrule
Shoulders + Upper Arms
& 0.843{\tiny$\pm$0.001}
& 72.8{\tiny$\pm$1.1}
& 49.0{\tiny$\pm$0.3}
& 0.0014{\tiny$\pm$0.0000}
& 0.001{\tiny$\pm$0.001}
& 51.2{\tiny$\pm$0.6}
& 50.4{\tiny$\pm$0.9}
& 75.3{\tiny$\pm$0.6}
& 61.7{\tiny$\pm$0.7} \\
\midrule
Forearms + Hands
& 0.840{\tiny$\pm$0.001}
& 68.6{\tiny$\pm$1.3}
& 42.7{\tiny$\pm$0.3}
& 0.0094{\tiny$\pm$0.0002}
& 0.013{\tiny$\pm$0.002}
& 51.9{\tiny$\pm$0.6}
& 46.6{\tiny$\pm$0.7}
& 69.6{\tiny$\pm$0.5}
& 57.3{\tiny$\pm$0.4} \\
\midrule
Pelvis + Upper Legs
& 0.797{\tiny$\pm$0.001}
& 90.5{\tiny$\pm$0.6}
& 39.1{\tiny$\pm$0.2}
& 0.0087{\tiny$\pm$0.0001}
& 0.012{\tiny$\pm$0.001}
& 51.8{\tiny$\pm$0.6}
& 37.1{\tiny$\pm$0.8}
& 64.9{\tiny$\pm$0.6}
& 50.0{\tiny$\pm$0.6} \\
\midrule
Lower Legs + Feet
& 0.801{\tiny$\pm$0.001}
& 77.7{\tiny$\pm$0.8}
& 44.0{\tiny$\pm$0.3}
& 0.0046{\tiny$\pm$0.0002}
& 0.006{\tiny$\pm$0.001}
& 53.2{\tiny$\pm$0.8}
& 28.3{\tiny$\pm$0.9}
& 51.4{\tiny$\pm$0.7}
& 39.4{\tiny$\pm$0.7} \\
\midrule
Head + Hands + Feet
& 0.848{\tiny$\pm$0.001}
& 65.9{\tiny$\pm$0.9}
& 47.6{\tiny$\pm$0.4}
& 0.0029{\tiny$\pm$0.0002}
& 0.004{\tiny$\pm$0.001}
& 51.1{\tiny$\pm$0.2}
& 53.0{\tiny$\pm$0.8}
& 74.2{\tiny$\pm$0.7}
& 62.7{\tiny$\pm$0.6} \\
\bottomrule
\end{tabular}%
}
\end{table}

\textbf{Correspondence across body locations.}
Applying the fixed AnyMo model to CIP-MTwAwinda yields paired cosine similarities of 0.775-0.848 and R@1 of 28.3\%-53.0\% across all seven configurations (\autoref{tab:cip_setup_alignment}).
Matching synchronized windows among 1,000 opposite-domain candidates demonstrates that the representation preserves motion-instance information across real and synthetic signals at diverse body locations, including the torso and lower limbs.
Adding foot observations to the head-and-hands configuration further raises R@1 from 48.9\% to 53.0\%, illustrating the complementary motion information integrated by the shared full-body representation.
Together with the downstream zero-shot HAR evaluation, these results support transfer across wearable sensing configurations without dataset-specific adaptation.

\subsubsection{Controlled Evaluation on Nymeria}

\begin{table}[h]
\caption{Controlled setup-transfer and real-synthetic alignment on Nymeria. Results are mean{\tiny$\pm$std} over 20 sampling seeds. Paired retrieval is evaluated within each activity's 100 candidate windows and averaged over both directions.}
\label{tab:nymeria_setup_alignment}
\centering
\scriptsize
\setlength{\tabcolsep}{1.2pt}
\renewcommand{\arraystretch}{1.05}
\resizebox{\linewidth}{!}{%
\begin{tabular}{@{}l|*{4}{c}|*{4}{c}|*{3}{c}@{}}
\toprule
Control / Variant
& \makecell[c]{Cos. Sim.$\,\uparrow$}
& \makecell[c]{Domain Classifier\\Acc. (\%)$\,\to 50$}
& \makecell[c]{kNN Domain\\Mixing (\%)$\,\to 50$}
& \makecell[c]{MMD$^2\,\downarrow$}
& \makecell[c]{Activity\\Silhouette$\,\uparrow$}
& \makecell[c]{Domain\\Silhouette$\,\to 0$}
& \makecell[c]{Activity\\Purity (\%)$\,\uparrow$}
& \makecell[c]{Domain\\Purity (\%)$\,\to 50$}
& \makecell[c]{Paired\\R@1 (\%)$\,\uparrow$}
& \makecell[c]{Paired\\R@5 (\%)$\,\uparrow$}
& \makecell[c]{Paired\\MRR (\%)$\,\uparrow$} \\
\midrule
\textbf{AnyMo}
& 0.856{\tiny$\pm$0.002}
& 72.8{\tiny$\pm$0.9}
& 45.8{\tiny$\pm$0.4}
& 0.0052{\tiny$\pm$0.0003}
& 0.057{\tiny$\pm$0.012}
& 0.006{\tiny$\pm$0.001}
& 54.1{\tiny$\pm$2.8}
& 53.9{\tiny$\pm$0.8}
& 74.7{\tiny$\pm$1.2}
& 87.7{\tiny$\pm$1.1}
& 80.7{\tiny$\pm$1.1} \\
\midrule
w/o geometry-aware simulation
& 0.507{\tiny$\pm$0.006}
& 98.0{\tiny$\pm$0.4}
& 3.2{\tiny$\pm$0.4}
& 0.2349{\tiny$\pm$0.0063}
& $-0.055${\tiny$\pm$0.009}
& 0.307{\tiny$\pm$0.009}
& 28.3{\tiny$\pm$1.5}
& 95.4{\tiny$\pm$1.4}
& 4.7{\tiny$\pm$0.6}
& 15.2{\tiny$\pm$1.1}
& 11.9{\tiny$\pm$0.7} \\
\midrule
w/o masked cross-view predictive contrastive
& 0.913{\tiny$\pm$0.002}
& 54.5{\tiny$\pm$0.8}
& 53.1{\tiny$\pm$0.3}
& 0.0019{\tiny$\pm$0.0002}
& $-0.042${\tiny$\pm$0.008}
& 0.002{\tiny$\pm$0.001}
& 35.8{\tiny$\pm$1.6}
& 52.9{\tiny$\pm$0.5}
& 96.5{\tiny$\pm$0.7}
& 98.8{\tiny$\pm$0.4}
& 97.5{\tiny$\pm$0.5} \\
\midrule
Orientation augmentation
& 0.855{\tiny$\pm$0.003}
& 71.8{\tiny$\pm$0.9}
& 46.2{\tiny$\pm$0.5}
& 0.0052{\tiny$\pm$0.0002}
& 0.056{\tiny$\pm$0.011}
& 0.006{\tiny$\pm$0.001}
& 53.4{\tiny$\pm$2.8}
& 54.2{\tiny$\pm$0.9}
& 74.4{\tiny$\pm$1.2}
& 87.3{\tiny$\pm$1.0}
& 80.4{\tiny$\pm$1.0} \\
\midrule
Head only
& 0.832{\tiny$\pm$0.003}
& 73.2{\tiny$\pm$0.8}
& 48.5{\tiny$\pm$0.4}
& 0.0030{\tiny$\pm$0.0002}
& $-0.038${\tiny$\pm$0.006}
& 0.003{\tiny$\pm$0.001}
& 38.7{\tiny$\pm$1.7}
& 53.2{\tiny$\pm$0.8}
& 65.6{\tiny$\pm$1.5}
& 82.6{\tiny$\pm$1.2}
& 73.4{\tiny$\pm$1.3} \\
\midrule
Two wrists only
& 0.848{\tiny$\pm$0.003}
& 70.2{\tiny$\pm$0.7}
& 44.6{\tiny$\pm$0.6}
& 0.0070{\tiny$\pm$0.0003}
& 0.041{\tiny$\pm$0.013}
& 0.009{\tiny$\pm$0.001}
& 52.0{\tiny$\pm$2.6}
& 54.2{\tiny$\pm$0.7}
& 72.7{\tiny$\pm$1.4}
& 86.1{\tiny$\pm$1.4}
& 78.9{\tiny$\pm$1.3} \\
\midrule
Device 2
& 0.858{\tiny$\pm$0.002}
& 72.9{\tiny$\pm$0.9}
& 46.2{\tiny$\pm$0.5}
& 0.0048{\tiny$\pm$0.0003}
& 0.056{\tiny$\pm$0.013}
& 0.006{\tiny$\pm$0.001}
& 52.9{\tiny$\pm$2.3}
& 53.5{\tiny$\pm$0.8}
& 75.5{\tiny$\pm$0.9}
& 87.6{\tiny$\pm$0.9}
& 81.1{\tiny$\pm$0.8} \\
\bottomrule
\end{tabular}%
}
\end{table}

\textbf{Evaluation protocol.}
We evaluate paired real and synthetic IMU embeddings using 1,000 synchronized motion windows from ten activity categories, with 100 windows per category.
Within each controlled comparison, we keep the participants, motion windows, activity labels, and preprocessing fixed.
For each window and observed body segment, we sample a synthetic body-surface candidate, thereby incorporating within-segment placement variation into the reference setting.
We then evaluate shared orientation perturbations applied to the paired real and synthetic inputs, head-only and two-wrist sensor subsets, and replacement of the real IMU with a synchronized, co-located second device.
Orientation perturbations comprise an in-plane rotation in $[-180^\circ,180^\circ]$ and local-axis tilts in $[-10^\circ,10^\circ]$.
All results are reported as mean$\pm$standard deviation over 20 sampling seeds.

\textbf{Metrics.}
Paired cosine similarity and bidirectional exact-pair retrieval measure whether real and synthetic observations of the same motion window remain close and distinguishable from other windows.
Nymeria retrieval uses 100 candidates within each activity, testing motion-instance correspondence among semantically similar examples.
Domain-classification accuracy closer to 50\%, nearest-neighbor domain mixing closer to 50\%, and lower MMD$^2$ indicate reduced domain distinguishability and greater distributional overlap.
Domain silhouette near zero and domain purity near 50\% indicate that clusters are less organized by signal source.
Higher activity silhouette and activity purity indicate clearer activity-level organization.

\textbf{Reference alignment and geometric modeling.}
Replacing geometry-aware surface simulation with fixed joint-mounted simulation markedly weakens real-synthetic correspondence (\autoref{tab:nymeria_setup_alignment}): paired R@1 falls from 74.7\% to 4.7\%, and cosine similarity decreases from 0.856 to 0.507.
Domain-classification accuracy rises from 72.8\% to 98.0\%, neighborhood mixing falls from 45.8\% to 3.2\%, and MMD$^2$ increases from 0.0052 to 0.2349, indicating that the resulting embeddings separate strongly by signal source.
Activity purity also drops from 54.1\% to 28.3\%, showing weaker organization by motion semantics.
AnyMo's geometry-aware simulation thus supports both cross-domain correspondence and activity structure across sampled within-segment placements, making synthetic pretraining more relevant to real wearable observations.

\textbf{Activity-structured representation learning.}
Replacing masked cross-view predictive-contrastive learning with full-view contrastive learning reduces activity silhouette from 0.057 to $-0.042$ and activity purity from 54.1\% to 35.8\%.
The full-view variant attains higher exact-pair R@1 (96.5\% versus 74.7\%), but its weaker activity structure shows that accurate instance matching alone does not ensure activity-level separation.
AnyMo combines positive activity silhouette and higher activity purity with strong paired correspondence, indicating that sparse-to-full predictive learning helps organize cross-setup motion instances into activity-discriminative representations.

\textbf{Orientation and device variation.}
Applying shared orientation perturbations to the real-synthetic pairs leaves R@1 nearly unchanged (74.7\% to 74.4\%) and preserves activity silhouette (0.057 to 0.056).
Replacing the real sensor with the synchronized second device likewise gives comparable R@1 (75.5\%), MRR (81.1\%), and activity purity (52.9\%) to the reference configuration (74.7\%, 80.7\%, and 54.1\%).
The preserved pairing and activity structure show that AnyMo retains motion information under the evaluated orientation and device changes, consistent with learning across varied wearable setups.

\textbf{Sparse sensor configurations.}
Reducing the head-and-wrists configuration to two wrists yields R@1 of 72.7\%, MRR of 78.9\%, and activity purity of 52.0\%, close to the reference values of 74.7\%, 80.7\%, and 54.1\%.
Using only the head lowers R@1 to 65.6\% and activity purity to 38.7\%, showing that wrist observations contribute information useful for both motion matching and activity separation.
The head-only configuration still retrieves corresponding motion windows with an MRR of 73.4\%.
These results show that AnyMo accommodates sparse sensor subsets while exploiting the complementary activity information available from additional body locations.

\section{AnyMo Bench}
\label{sec:anymo_bench}
Beyond model training, AnyMo-180 also enables a controlled benchmark construction.
We further derive \textbf{AnyMo Bench}, an in-the-wild HAR benchmark from real Nymeria IMU streams~\citep{ma2024nymeria}.
The benchmark is designed to stress two forms of generalization that are central to wearable sensing in realistic deployments: recognizing fine-grained daily activities on unseen subjects, and transferring across different IMU units mounted at the same body position.
Such setup shifts are a persistent challenge for wearable HAR, where activity recognition can be affected by inter-subject motion variation, device hardware, exact body placement, orientation, and collection protocol~\citep{haresamudram2025past, cai2025towards, bian2026foundation}.
Nymeria enables this evaluation because it contains synchronized real IMU streams from three body positions: Head, Left Wrist, and Right Wrist with two co-located IMUs per position. In the cross-device setting, models train on the first IMU at each position and are tested on the second IMU at the same position, so the body placement is fixed while the IMU unit changes.

\textbf{Evaluation settings.}
We use a subject-disjoint 8:2 split with seed 42, yielding 157 training subjects and 39 test subjects from 196 subjects. AnyMo Bench contains 154{,}839 activity windows and covers 211.6 hours of real in-the-wild IMU data. The split contains 123{,}874 training windows and 30{,}965 test windows for the eligible Fine150 label space, with no class missing from either split. We report four settings: \textit{Fine150 / Unseen Subject}, \textit{Fine150 / Unseen Subject + Cross Device}, \textit{Core50 / Unseen Subject}, and \textit{Core50 / Unseen Subject + Cross Device}. In the Unseen Subject settings, train and test examples use the first IMU at each body position. In the Unseen Subject + Cross Device settings, training uses the first IMU, but testing uses the second co-located IMU for the held-out subjects.

\textbf{Label spaces.}
Fine150 is a fine-grained 150-class label space derived from AnyMo-180. Starting from the 180 curated classes, we remove labels that are unstable under the subject split, including classes with fewer than two test subjects or fewer than 10 test windows, and exclude a small number of highly context-dependent labels whose distinction depends heavily on visual or semantic context rather than IMU-observable motion. Fine150 remains long-tailed, with per-class counts ranging from 47 to 19{,}715 rows, a median of 412.5, and a mean of 1032.3. Core50 is a coarser 50-class label space (still more fine-grained than most existing IMU-based HAR datasets) built by merging Fine150 labels with IMU-pose-aware semantics: labels are combined when they share similar body-motion signatures, are low-sample, and remain interpretable as a coherent core activity. Large, motion-distinct, or already reliable classes are kept separate. After aggregation, Core50 class counts range from 72 to 19{,}715 rows, with a median of 1923.5 and a mean of 3096.8.

\begin{table}[h]
\caption{IMU-based HAR baseline results on AnyMo Bench.}
\label{tab:anymo_bench_baselines}
\centering
\begin{tabular}{@{}lccc@{}}
\toprule
Model & Acc@1 & Acc@5 & Macro-F1 \\
\midrule
\multicolumn{4}{@{}l}{\cellcolor{gray!12}\textit{Fine150 / Unseen Subject}} \\
DeepConvLSTM & 35.3 & 63.0 & \underline{17.2} \\
MantisV2 & \textbf{38.5} & \textbf{65.2} & \textbf{22.8} \\
COMODO & \underline{37.8} & \underline{65.2} & 16.0 \\
\midrule
\multicolumn{4}{@{}l}{\cellcolor{gray!12}\textit{Fine150 / Unseen Subject + Cross Device}} \\
DeepConvLSTM & 1.9 & 9.5 & 0.4 \\
MantisV2 & \underline{14.4} & \underline{39.7} & \textbf{8.6} \\
COMODO & \textbf{24.0} & \textbf{50.6} & \underline{8.0} \\
\midrule
\multicolumn{4}{@{}l}{\cellcolor{gray!12}\textit{Core50 / Unseen Subject}} \\
DeepConvLSTM & 43.2 & 75.4 & 34.5 \\
MantisV2 & \underline{45.8} & \underline{76.8} & \textbf{41.3} \\
COMODO & \textbf{46.2} & \textbf{78.8} & \underline{37.3} \\
\midrule
\multicolumn{4}{@{}l}{\cellcolor{gray!12}\textit{Core50 / Unseen Subject + Cross Device}} \\
DeepConvLSTM & 1.8 & 12.1 & 0.6 \\
MantisV2 & \underline{16.6} & \underline{48.7} & \underline{18.4} \\
COMODO & \textbf{32.6} & \textbf{67.8} & \textbf{23.3} \\
\bottomrule
\end{tabular}
\end{table}

\textbf{Baselines and metrics.}
We evaluate three baselines. DeepConvLSTM is a classic IMU-based HAR method~\citep{ordonez2016deep}, and MantisV2 is one of the best recent time-series classification foundation models~\citep{feofanov2026mantisv2}; both supervised IMU baselines are trained for 100 epochs on the AnyMo Bench training split. We also adapt COMODO, a self-supervised multimodal baseline for video-to-IMU representation learning~\citep{chen2025comodo}, using MantisV2 as the IMU backbone and TimeSformer as the video backbone, and train it for 20 epochs.
AnyMo Bench shares the Nymeria source domain and underlying recordings
used for AnyMo's synthetic pretraining; we therefore report other
baselines here rather than treating it as an independent zero-shot
benchmark for AnyMo.
We synchronize all IMU streams to a common 60 Hz temporal grid to align the two co-located Nymeria IMU units at each body position, while preserving temporal detail that supports the fine-grained daily-activity taxonomy~\citep{straczkiewicz2021systematic}. The resulting activity windows are variable length, with durations concentrated around 5 seconds. Across the eligible Fine150/Core50 windows, 132{,}915 of 154{,}839 windows (85.84\%) contain 300 IMU timesteps, with lengths ranging from 60 to 300 timesteps (1.0-5.0 s). For fixed-length baselines such as DeepConvLSTM, windows shorter than 300 timesteps are zero-padded and longer windows are split into non-overlapping 300-timestep segments, dropping any remaining tail. For baselines with model-specific variable-length handling, such as MantisV2 and COMODO with a MantisV2 IMU backbone, IMU segments are resized to the model input length by interpolation.
We report Acc@1, Acc@5, and macro-F1 to measure exact recognition, near-miss retrieval among many candidate activities, and class-balanced performance.

\textbf{Results and implications.}
As shown in \autoref{tab:anymo_bench_baselines}, the results indicate that AnyMo Bench is challenging. On Fine150 / Unseen Subject, MantisV2 and COMODO reach comparable top-line accuracy, with MantisV2 at 38.5\% Acc@1 and 65.2\% Acc@5 and COMODO at 37.8\% Acc@1 and 65.2\% Acc@5, while DeepConvLSTM reaches 35.3\% Acc@1 and 63.0\% Acc@5. These results are meaningful for a 150-class in-the-wild unseen-subject task, but the macro-F1 scores indicate substantial remaining difficulty on tail and fine-grained classes. Core50 improves recognition accuracy, with COMODO reaching 46.2\% Acc@1 and 78.8\% Acc@5, while MantisV2 obtains the strongest macro-F1 at 41.3\%; however, the remaining error shows that reducing the label granularity does not make the task trivial. Cross-device evaluation remains much harder: COMODO gives the strongest cross-device Acc@1, reaching 24.0\% on Fine150 and 32.6\% on Core50, but the gap from same-device unseen-subject recognition remains large, revealing substantial room for same-position different-device transfer.

These results suggest that the difficulty comes not only from label granularity, but also from realistic subject and device shifts. Because the benchmark construction combines automatic label proposal, embedding-based semantic label consolidation, human-expert review, enum-label assignment, full class-wise auditing, relabel/drop decisions, and IMU-pose-aware label-space aggregation, AnyMo Bench provides a challenging, in-the-wild, and carefully curated testbed for future work on robust wearable motion recognition.

\section{Downstream Evaluation Dataset Details}
\label{sec:downstream_evaluation_dataset}
The zero-shot recognition benchmark contains 14 downstream datasets that are never used for training AnyMo.
\autoref{tab:downstream_dataset_details} lists each dataset's activity classes and sensor placements.
Sensor placements are reported using readable AnyMo graph segment names exposed to the model during evaluation, after mapping each dataset sensor to the closest node(s) in AnyMo's 23-node body graph.
For EgoExo4D, Ego4D, and MMEA, we follow the train/test splits from COMODO~\citep{chen2025comodo}\footnote{\url{https://huggingface.co/CRUISEResearchGroup/COMODO-AnyMo-Bench}}.
For OpenPack, we use the OpenPack Challenge 2022 split~\citep{yoshimura2024openpack}\footnote{\url{https://open-pack.github.io}} and discard contiguous operation segments shorter than 1 s or longer than 30 s; this removes 204 of 10{,}512 valid labeled segments (1.94\%) across the official split.
For the remaining datasets, we use the train/test splits distributed with UniMTS~\citep{zhang2024unimts}\footnote{\url{https://huggingface.co/xiyuanz/UniMTS}}.

{
\setlength{\tabcolsep}{2.5pt}
\setlength{\LTcapwidth}{\linewidth}
\renewcommand{\arraystretch}{1.08}
\begin{longtable}{@{}C{0.12\textwidth}C{0.12\textwidth}L{0.48\textwidth}C{0.23\textwidth}@{}}
\caption{Downstream zero-shot HAR datasets, activity classes, and sensor placements.}
\label{tab:downstream_dataset_details}\\
\toprule
Dataset & \#~Classes & Classes & Sensor Placements \\
\midrule
\endfirsthead
\toprule
Dataset & \#~Classes & Classes & Sensor Placements \\
\midrule
\endhead
\bottomrule
\endfoot
Opportunity & 4 & stand, walk, sit, lie & L3, Right Upper Arm, Right Forearm, Left Upper Arm, Left Forearm \\
\midrule
UCI-HAR & 6 & walk, walk upstairs, walk downstairs, sit, stand, lay & L5 \\
\midrule
w-HAR & 7 & walk, sit, stand, jump, lie down, stairs up, stairs down & Right Foot \\
\midrule
RealWorld & 8 & lying, jumping, standing, walking, sitting, climbing down, climbing up, running & Left Shoulder, Left Forearm, Head, Left Foot, Left Upper Leg, Left Upper Arm, L5 \\
\midrule
TNDA-HAR & 8 & sitting, standing, lying down, ascending stairs, descending stairs, riding, walking, jogging & Right Forearm, Left Lower Leg, Right Hand, Left Foot, T8 \\
\midrule
EgoExo4D & 8 & Basketball, Bike Repair, Cooking, Dance, Health, Music, Rock Climbing, Soccer & Head \\
\midrule
OpenPack & 10 & Picking, Relocate Item Label, Assemble Box, Insert Items, Close Box, Attach Box Label, Scan Label, Attach Shipping Label, Put on Back Table, Fill out Order & Right Forearm, Left Forearm, Right Upper Arm, Left Upper Arm \\
% \midrule
PAMAP2 & 12 & lying, sitting, standing, walking, running, cycling, nordic walking, ascending stairs, descending stairs, vacuum cleaning, ironing, rope jumping & Right Hand, T8, Right Foot \\
\midrule
USC-HAD & 12 & walk forward, walk left, walk right, walk upstairs, walk downstairs, run forward, jump up, sit, stand, sleep, elevator up, elevator down & Right Upper Leg \\
\midrule
WISDM & 18 & walking, jogging, stairs, sitting, standing, typing, brushing teeth, eating soup, eating chips, eating pasta, drinking cup, eating sandwich, kicking ball, playing ball, dribbling ball, writing, clapping, folding clothes & Right Hand \\
\midrule
DSADS & 19 & sitting, standing, lying on back, lying on right side, ascending stairs, descending stairs, standing in an elevator still, moving around in an elevator, walking slowly, walking on a treadmill in flat positions, walking on a treadmill in inclined positions, running on a treadmill fast, exercising on a stepper, exercising on a cross trainer, cycling on an exercise bike in horizontal positions, cycling on an exercise bike in vertical positions, rowing, jumping, playing basketball & T8, Right Hand, Left Hand, Right Lower Leg, Left Lower Leg \\
\midrule
UTD-MHAD & 27 & right arm swipe to the left, right arm swipe to the right, right hand wave, two hand front clap, right arm throw, cross arms in the chest, basketball shoot, right hand draw x, right hand draw circle (clockwise), right hand draw circle (counter clockwise), draw triangle, bowling (right hand), front boxing, baseball swing from right, tennis right hand forehand swing, arm curl (two arms), tennis serve, two hand push, right hand knock on door, right hand catch an object, right hand pick up and throw, jogging in place, walking in place, sit to stand, stand to sit, forward lunge (left foot forward), squat (two arms stretch out) & Right Hand for labels 0-20; Right Upper Leg for labels 21-26 \\
\midrule
Ego4D & 31 & baker, bike, bike mechanic, biology experiments, car / commuting / road trip, car / scooter washing, carpenter, cleaning / laundry, cooking, crafting / knitting / sewing / drawing / painting, cycling / jogging, eating, farmer, fixing something in the home, gardening, household management / caring for kids, indoor navigation (walking), construction / renovation jobs, playing board games, playing cards, playing games / video games, playing with pets, potting plants (indoor), practicing a musical instrument, reading books, scooter mechanic, walking on street, watching tv, working at desk, working out at home, working out outside & Head \\
% \midrule
MMEA & 32 & upstairs, downstairs, drinking, fall, reading, sweep floor, cut fruits, mop floor, writing, wipe table, wash hand, standing, play phone, type pc, eating, cooking, pick up phone, drop trash, fold clothes, walking, play card, brush teeth, wash dish, moving sth, type phone, chat, open close door, ride bike, sit stand, take drop sth, shopping, watch TV & Head \\
\end{longtable}
}

\section{Baseline Details}
\label{sec:baseline_details}
For activity recognition, all baselines are evaluated with the same dataset-specific candidate activity vocabularies used by AnyMo.
For ImageBind~\citep{girdhar2023imagebind} and IMU2CLIP~\citep{moon2023imu2clip}, we use the checkpoints pre-trained on Ego4D for all non-Ego4D recognition datasets.
For Ego4D evaluation, we instead pre-train each baseline on MMEA and evaluate the resulting checkpoint on Ego4D, avoiding an Ego4D-pre-trained checkpoint on the Ego4D target dataset.
Both ImageBind and IMU2CLIP use the same 60 Hz temporal setting as AnyMo.
For IMUGPT~\citep{leng2023generating}, we follow the released default configuration and use its 20 Hz IMU input setting.
For HARGPT~\citep{ji2024hargpt}, we follow the default prompt-based setting and feed IMU at 10 Hz.
For UniMTS~\citep{zhang2024unimts}, we use the released checkpoint generated and pre-trained at 20 Hz, and evaluate it under the same 20 Hz setting.
For NormWear~\citep{luo2026toward}, we follow the official implementation and its 65 Hz input setting.
For the Gemma 4 26B text and plot baselines~\citep{googledeepmind2026gemma4modelcard}, we downsample IMU inputs to 10 Hz; even at 10 Hz, we use a 65{,}536-token context window so that all benchmark datasets can be processed without truncation.
For motion captioning, GPT-5.4 Mini and Gemma 4 26B are prompted with raw numerical IMU signals and sensor-location context to generate natural-language motion descriptions without task-specific training.
For EgoLM~\citep{hong2025egolm}, we adapt its architecture to raw IMU captioning and train it using the same training data and protocol as AnyMo.

\section{Prompt Analysis}
\subsection{Prompt Sensitivity}
\label{sec:prompt_sensitivity}
Zero-shot recognition embeds each candidate activity label through the language backbone, so the text-side representation is sensitive to how the label is prompted.
To quantify this effect, we evaluate four fixed prompt formats with the same recognition protocol as \autoref{tab:zeroshot_har}.
The fixed prompts are: \textit{bare label}, which uses the class name directly; \textit{person prompt}, ``a person is \{label\}''; \textit{IMU prompt}, ``wearable IMU motion of \{label\}''; and \textit{activity prompt}, ``the activity is \{label\}''.
We compare them with the final AnyMo setting, which replaces manual text templates with an 8-token learnable soft prompt on the text/label branch.

\begin{table}[h]
\caption{Prompt sensitivity for zero-shot HAR. Results averaged over 14 downstream datasets.}
\label{tab:prompt_sensitivity}
\centering
\begin{tabular}{lccc}
\toprule
Text-side prompt & Acc & F1 & R@2 \\
\midrule
Bare label & 33.1 & 25.9 & 54.3 \\
Person prompt & 35.1 & 28.4 & 54.8 \\
IMU prompt & 28.3 & 22.5 & 47.0 \\
Activity prompt & 35.2 & 28.0 & 56.1 \\
Learnable prompt (AnyMo) & \textbf{35.7} & \textbf{29.5} & \textbf{57.5} \\
\bottomrule
\end{tabular}
\end{table}

The fixed prompts produce noticeably different rankings even though the label vocabulary, IMU embeddings, and evaluation protocol are unchanged.
For example, the IMU-centric template is weaker than the person- and activity-centric templates on this benchmark, indicating that manually written label contexts can shift the text embeddings in ways that are not consistently beneficial for recognition.
The learnable soft prompt avoids committing to a hand-written template: during contrastive instruction tuning, it learns a task-specific textual context for activity labels that better matches the IMU embedding space.
This also removes the need to choose a dataset-specific prompt template at evaluation time.
Accordingly, AnyMo uses the learnable prompt in all final zero-shot recognition results, where it gives the best average Acc, F1, and Recall@2 among the compared single-prompt settings.

\subsection{Prompt Templates}
\label{sec:prompt_templates}
\autoref{fig:prompt_templates} summarizes the text templates used by AnyMo.
The IMU-token language-model pre-training stage is trained to generate IMU token sequences.
The narration and MCQ instruction templates are used during language-modeling instruction tuning; captioning evaluation uses the narration template, and MCQ-style evaluation uses the MCQ template.
The contrastive IMU and text templates are used during contrastive instruction tuning and are used for retrieval and zero-shot recognition, with the learnable soft prompt prepended to the text/label branch in the final AnyMo checkpoint.
\newcommand{\prompttemplatebox}[2]{%
\noindent\fbox{%
\begin{minipage}{0.94\linewidth}
\colorbox{black!75}{\parbox{\dimexpr\linewidth-2\fboxsep\relax}{\strut\color{white}\bfseries #1}}
\par\vspace{0.6em}
#2
\par\vspace{0.5em}
\end{minipage}}}

\begin{figure*}[h]
\centering
\small
\prompttemplatebox{IMU-Token LM Pre-Training}{
\texttt{\{IMU\_TOKEN\_SEQUENCE\}}
}

\vspace{0.8em}
\prompttemplatebox{Narration Instruction Tuning / Captioning Evaluation}{
Describe the human motion represented by the wearable IMU motion tokens.\newline
The IMU tokens are from IMU sensors attached to the user's \texttt{\{SENSOR\_CONTEXT\}}.\newline
Input IMU token:\newline
\texttt{\{IMU\_TOKEN\_SEQUENCE\}}\newline
\textbf{Assistant target:} \texttt{\{MOTION\_NARRATION\}}
}

\vspace{0.8em}
\prompttemplatebox{MCQ Instruction Tuning / MCQ Evaluation}{
Recognize the activity represented by the wearable IMU motion tokens.\newline
The IMU tokens are from IMU sensors attached to the user's \texttt{\{SENSOR\_CONTEXT\}}.\newline
Input IMU token:\newline
\texttt{\{IMU\_TOKEN\_SEQUENCE\}}\newline
CHOICES:\newline
\texttt{\{CHOICE\_KEY\}: \{ACTIVITY\_LABEL\}}\newline
Choose the best matching option. Output the option key followed by the selected activity label.\newline
\textbf{Assistant target:} \texttt{\{ANSWER\_KEY\}: \{ACTIVITY\_LABEL\}}
}

\vspace{0.8em}
\prompttemplatebox{Contrastive IMU Prompt}{
Represent the human motion from the wearable IMU motion tokens.\newline
The IMU tokens are from IMU sensors attached to the user's \texttt{\{SENSOR\_CONTEXT\}}.\newline
Input IMU token:\newline
\texttt{\{IMU\_TOKEN\_SEQUENCE\}}\newline
Return a compact embedding of the motion.
}

\vspace{0.8em}
\prompttemplatebox{Contrastive Text / Label Prompt}{
Represent the human motion described by the text.\newline
Motion description:\newline
\texttt{\{NARRATION\_OR\_ACTIVITY\_LABEL\}}\newline
Return a compact embedding of the motion.
}
\caption{Prompt templates used for motion-language pre-training, instruction tuning, contrastive tuning, and evaluation.}
\label{fig:prompt_templates}
\end{figure*}

\input{sec/2_related_works}

% \newpage

% \section{Technical appendices and supplementary material}
% Technical appendices with additional results, figures, graphs, and proofs may be submitted with the paper submission before the full submission deadline (see above). You can upload a ZIP file for videos or code, but do not upload a separate PDF file for the appendix. There is no page limit for the technical appendices. 

% Note: Think of the appendix as ``optional reading'' for reviewers. The paper must be able to stand alone without the appendix; for example, adding critical experiments that support the main claims to an appendix is inappropriate. 

%%%%%%%%%%%%%%%%%%%%%%%%%%%%%%%%%%%%%%%%%%%%%%%%%%%%%%%%%%%%

% \newpage
% \input{checklist.tex}

\end{document}

%% file: sec/1_introduction.tex
% !TEX root = ../neurips_2026.tex
\section{Introduction}
\label{sec:introduction}
Human motion is one of the most immediate expressions of human context in everyday life~\citep{rosenhahn2008human, aggarwal2011human}.
When people walk, cook, exercise, or interact with others or objects, their movement directly reflects their engagement with their surroundings over time.
Understanding this context is important for future proactive AI and human-centered computing systems, which must proactively respond to changing user contexts and adapt in real environments rather than simply waiting for explicit commands~\citep{schilit1994context, dey2001understanding}. 
The growing ubiquity of wearable and mobile devices, from watches and phones to earbuds, smart rings, AR glasses, and body-worn sensors, creates new opportunities for sensing human motion in the wild, thus developing context-aware AI systems~\citep{bulling2014tutorial, haresamudram2025past, cai2025towards}.

Yet sensing motion is not the same as understanding it. Inertial measurement unit (IMU) signals are semantically ambiguous: similar inertial patterns can arise from different activities depending on who is moving, how, and in what context. Resolving this ambiguity requires knowledge beyond closed-set activity labels. Language provides a natural source of such knowledge, as it is grounded in human descriptions of everyday behavior and supports compositional, open-ended semantics. Connecting wearable sensing to language therefore helps models interpret motion in terms that generalize beyond fixed labels, making it central to generalist wearable motion understanding.

However, wearable IMU signals are tightly coupled to how and where the device is worn, making robust modeling difficult across sensing setups~\citep{cai2025towards, kunze2014sensor, chang2020systematic, sztyler2016body}. A wrist-worn watch emphasizes arm motion, glasses capture head motion, and a phone in a pocket measures dynamics coupled to the torso and legs, even when the underlying activity is the same. 
Small changes in mounting position or orientation within a body part can further alter the measured acceleration and angular velocity, while device hardware and sampling introduce additional shifts~\citep{stisen2015smart,hong2024crosshar}. Consequently, models trained for one setup often struggle to transfer to another setup across users, devices, and datasets. 

\begin{figure}[t]
% \vspace{-1em}
    \centering
    \includegraphics[width=1\linewidth]{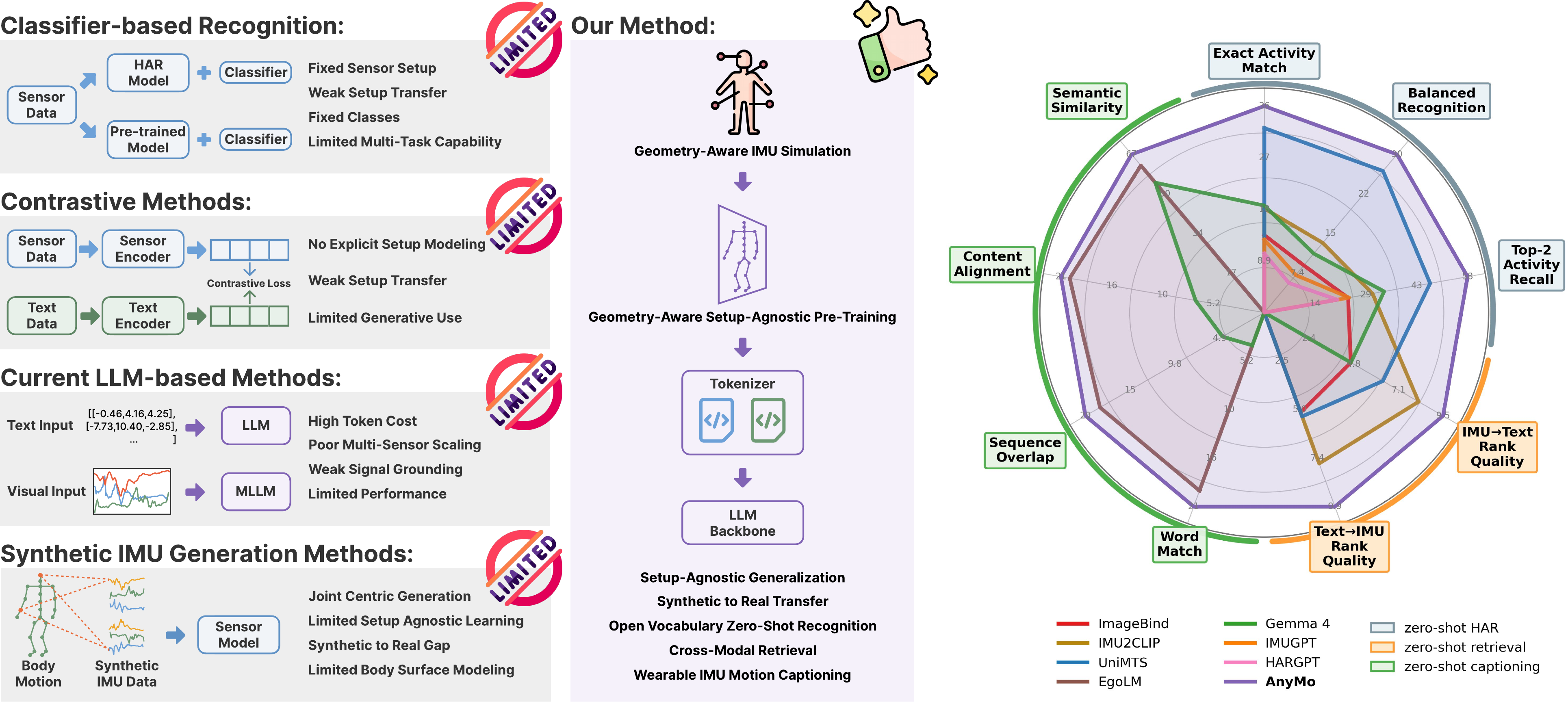}
    \caption{Method families for wearable human motion understanding and radar plot comparing the performance of AnyMo with baselines across various tasks and capabilities.}
    \label{fig:method_families}
    % \vspace{-2em}
\end{figure}

This setup dependence, combined with the challenges of grounding IMU in language, makes building a broadly useful wearable motion model difficult along three coupled axes~\citep{haresamudram2025past, cai2025towards, bian2026foundation}. 
% \ding{182} Large-scale, setup-diverse real IMU data are difficult to collect and remain fragmented across body placements, device hardware, sampling rates, users, and datasets, making it difficult to learn representations that remain stable under realistic setup shifts.
% \ding{183} Supervision is often scarce and semantically thin, since many wearable datasets provide only a small closed set of coarse activity labels rather than rich descriptions of how motion unfolds.
\ding{182} \textbf{Data and Supervision Scarcity}: Real IMU data is difficult to collect at scale. It remains fragmented across body placements, device hardware, sampling rates, and datasets, and supervision is often limited to a small closed set of coarse activity labels rather than rich descriptions of how motion unfolds.
\ding{183} \textbf{Limited Realism of Synthetic Augmentation}: Synthetic or augmented sensor data must expand setup coverage without losing physical realism, but existing generation pipelines often remain tied to specific labels, activities, or sparse sensor placements.
\ding{184} \textbf{Modality Gap}: Connecting IMU to language requires bridging continuous, multi-sensor motion signals with discrete textual concepts, a modality gap that direct prompting or simple contrastive alignment does not fully resolve and that grows more severe as the number of sensors, channels, and body locations increases. 
\autoref{fig:method_families} contextualizes these issues across classifier-based, contrastive, LLM-based, and synthetic-generation methods, which address parts of the problem but remain limited along different axes.

These challenges motivate our key insight: \textbf{wearable setup variation is structured rather than arbitrary}.
An IMU is attached to a body surface, and its signal is produced by the interaction of body motion, surface geometry, sensor orientation, and device response.
This structure provides a geometry- and physics-based inductive bias for learning setup-robust body motion representations.
We further argue that \textbf{language should be connected to wearable sensing through a compact motion representation rather than raw IMU streams}. 
Language models provide priors for open-vocabulary motion understanding, but raw numerical IMU tokens scale with sensors, channels, and time, while sensor location-specific tokenizers tie representations to fixed setups. Compact full-body tokens avoid both limitations, providing a stable interface between motion and language.

With these insights, we introduce \textbf{AnyMo}, a geometry-aware framework for setup-agnostic human motion modeling in the wild.
It aims to learn robust IMU representations across \textbf{Any} wearable setup for human \textbf{Mo}tion understanding.
As shown in \autoref{fig:method_families}, AnyMo connects \textbf{physics-grounded geometry-aware IMU simulation} with \textbf{geometry-aware setup-agnostic pre-training}, \textbf{full-body tokenization}, and \textbf{motion-language alignment}.
The simulation stage generates dense geometry-aware IMU candidates over body-surface placements, providing a broad and plausible distribution of wearable locations and orientations.
The pre-training stage constructs paired placement views and masked wearable observations, encouraging the encoder to learn setup-agnostic motion representations.
The tokenizer converts multi-position IMU observations into compact full-body motion tokens, aligned with an LLM for open-vocabulary recognition, cross-modal retrieval, and motion captioning.

To validate AnyMo, we benchmark it against state-of-the-art methods on three complementary tasks: zero-shot activity recognition, unseen cross-modal retrieval, and wearable IMU motion captioning.
For zero-shot recognition, we use 14 completely unseen downstream datasets spanning classic HAR benchmarks and in-the-wild settings.
Retrieval and captioning are evaluated under sim-to-real transfer on unseen Nymeria subjects and out-of-domain (OOD) zero-shot transfer to EgoExo4D.
Across all tasks, AnyMo shows significant gains over baselines, establishing it as a generalist model for wearable motion understanding. \autoref{tab:conceptual_comparison} summarizes AnyMo's setup-transfer
capabilities and supported motion-understanding tasks relative to prior methods. Our contributions are as follows:

\ding{182} \textbf{We propose physics-grounded, geometry-aware IMU simulation} over dense body-surface placements, providing diverse and plausible synthetic signals that bridge synthetic pre-training and real wearable IMU.
\ding{183} \textbf{We develop a setup-agnostic representation learning method and full-body IMU tokenization}, aligning motion across synthetic placement views and mapping multi-position IMU into full-body motion tokens.
\ding{184} \textbf{We introduce an IMU-language generalist model} that supports a variety of wearable motion understanding tasks.

\begin{table}[h]
\caption{Comparison with representative sensor-based models.
\cmpyes{} denotes support, \cmppartial{} partial support or release,
and \cmpno{} a capability not demonstrated or an official implementation
not found.
Placement and sensor-set transfer concern deployment under changed
placements/orientations and sensor combinations without setup-specific
retraining. Coord.-free inference indicates that within-segment sensor coordinates are not required at inference time. Related work is provided in
Appendix~\ref{sec:related_work}.}
\label{tab:conceptual_comparison}
\centering
\scriptsize
\setlength{\tabcolsep}{3pt}
\renewcommand{\arraystretch}{1}
\resizebox{\linewidth}{!}{%
\begin{tabular}{@{}lcccccccc@{}}
\toprule
\textbf{Method}
& \shortstack{\textbf{Coord.-free}\\\textbf{Inference}}
& \shortstack{\textbf{Open}\\\textbf{Source}}
& \shortstack{\textbf{Sensor-set}\\\textbf{Transfer}}
& \shortstack{\textbf{Surface}\\\textbf{Sim.}}
& \shortstack{\textbf{Placement}\\\textbf{Transfer}}
& \shortstack{\textbf{Zero-shot}\\\textbf{HAR}}
& \textbf{Retrieval}
& \textbf{Captioning} \\
\midrule

MoPFormer~\citep{zhang2025mopformer}
& \cmpyes & \cmpno & \cmpno & \cmpno
& \cmpno & \cmpno & \cmpno & \cmpno \\

SensorLLM~\citep{li2025sensorllm}
& \cmpyes & \cmpyes & \cmpno & \cmpno
& \cmpno & \cmpno & \cmpno & \cmpno \\

W2W~\citep{ray2025w2w}
& \cmpyes & \cmpno & \cmpno & \cmpyes
& \cmpno & \cmpno & \cmpno & \cmpno \\

EgoLM~\citep{hong2025egolm}
& \cmpyes & \cmpno & \cmpno & \cmpno
& \cmpno & \cmpno & \cmpno & \cmpyes \\

CrossHAR~\citep{hong2024crosshar}
& \cmpyes & \cmpyes & \cmpno & \cmpno
& \cmpno & \cmppartial & \cmpno & \cmpno \\

IMUGPT / 2.0~\citep{leng2023generating,leng2024imugpt2}
& \cmpyes & \cmpyes & \cmpno & \cmpno
& \cmpno & \cmppartial & \cmpno & \cmpno \\

Wonderwall~\citep{miao2026wonderwall}
& \cmpyes & \cmpno & \cmpyes & \cmpno
& \cmppartial & \cmpyes & \cmpno & \cmpno \\

IMU2CLIP~\citep{moon2023imu2clip}
& \cmpyes & \cmpyes & \cmpno & \cmpno
& \cmpno & \cmpyes & \cmpyes & \cmpno \\

IMUCoCo~\citep{zhou2025imucoco}
& \cmpno & \cmpyes & \cmpyes & \cmpyes
& \cmpyes & \cmpno & \cmpno & \cmpno \\

SensorLM~\citep{zhang2025sensorlm}
& \cmpyes & \cmppartial & \cmpno & \cmpno
& \cmpno & \cmpyes & \cmpyes & \cmpyes \\

UniMTS~\citep{zhang2024unimts}
& \cmpyes & \cmpyes & \cmpyes & \cmpno
& \cmppartial & \cmpyes & \cmpyes & \cmpno \\

\midrule
\textbf{AnyMo (ours)}
& \cmpyes & \cmpyes & \cmpyes & \cmpyes
& \cmpyes & \cmpyes & \cmpyes & \cmpyes \\
\bottomrule
\end{tabular}%
}
\end{table}

%% file: sec/3_method.tex
% !TEX root = ../neurips_2026.tex
\section{Methodology}
\label{sec:methodology}

\begin{wrapfigure}{r}{0.375\linewidth}
    % \vspace{-1em}
    \centering
    \includegraphics[width=\linewidth]{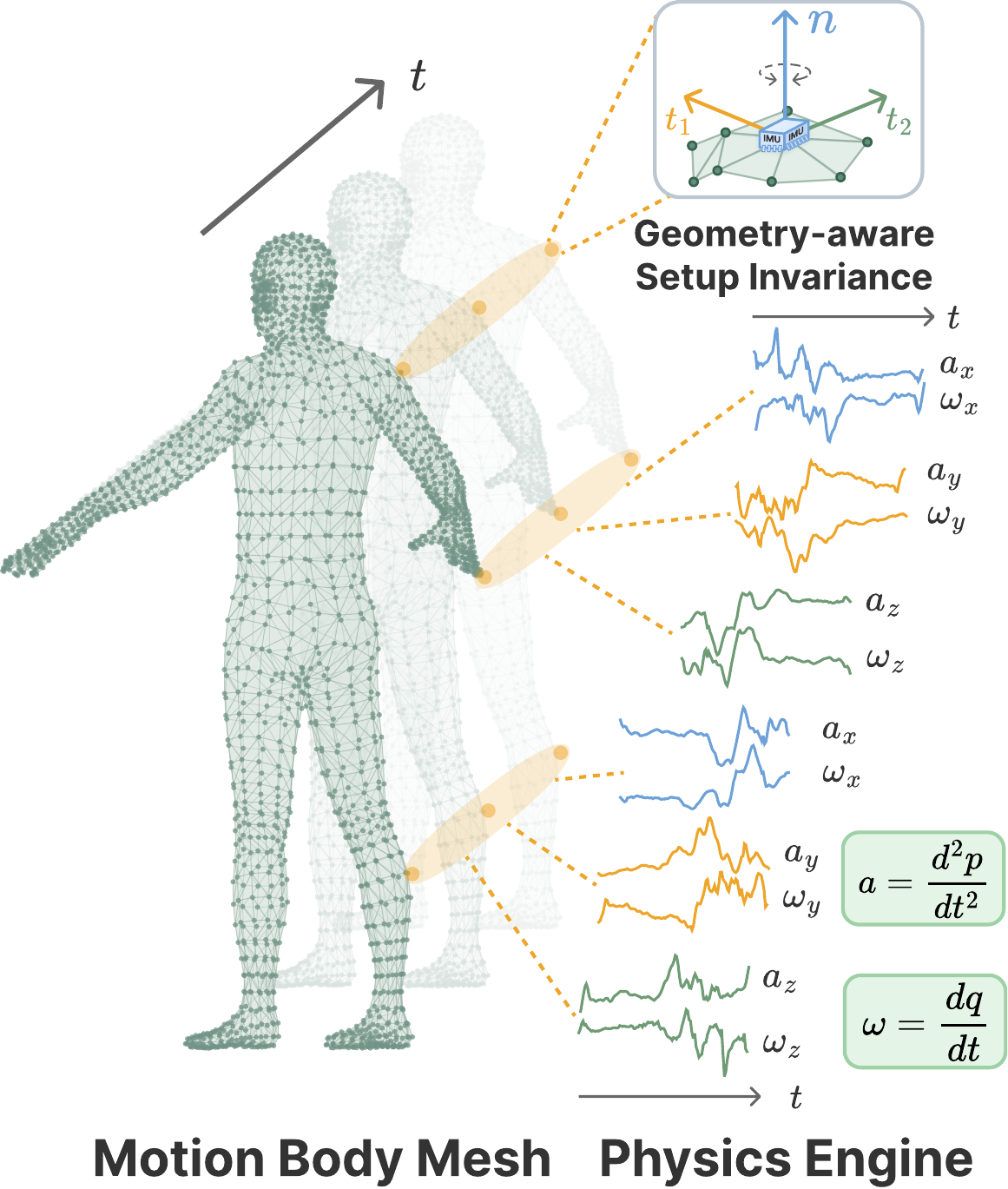}
    \caption{Physics-grounded geometry-aware motion simulation.}
    \label{fig:simulation}
    % \vspace{-0.8em}
\end{wrapfigure}

Building on the observation that wearable setup variation is structured, AnyMo targets wearable IMU motion understanding under variable sensing setups.
% In practice, a motion window may be observed by only a few wearable IMUs, and the measured signals depend on body locations, surface placements, sensor orientations, hardware response, and sampling protocol.
We follow the Nymeria~\citep{ma2024nymeria} body model and organize motion over $N_{\mathrm{seg}}=23$ anatomical segments.
We denote an IMU window as $\mathbf{x}\in\mathbb{R}^{T\times 6}$, where the last dimension contains three-axis acceleration and three-axis angular velocity.
The same motion can therefore yield different IMU windows across different wearable setups, whereas a real device typically provides only a partial observation of the body.
AnyMo aims to learn a representation that absorbs such partial, setup-specific IMU windows and preserves motion information that is useful across sensing setups and language-based tasks.
AnyMo combines geometry-aware IMU simulation (\autoref{sec:method_simulation}, \autoref{fig:simulation}), setup-agnostic representation learning and full-body tokenization (\autoref{sec:method_pretraining}, \autoref{fig:pretraining}), and motion-language alignment (\autoref{sec:method_motion_language}, \autoref{fig:cit_inference}).

\subsection{Physics-Grounded Geometry-Aware Motion Simulation}
\label{sec:method_simulation}
Motion skeleton and mesh data describe human body motion through segment positions, orientations, and posed body-surface geometry over time~\citep{ma2024nymeria}.
Wearable IMUs, however, measure local acceleration and angular velocity rather than body pose directly.
We synthesize wearable IMU windows by applying wearable IMU motion equations~\citep{oishi2025wimusim} to synchronized Nymeria body motion.
Unlike joint-centric simulation~\citep{zhang2024unimts, leng2023generating, leng2024imugpt2, wei2025one, miao2026wonderwall}, our goal is to model plausible wearable locations on the body surface, together with their local sensor frames and device noise.
For each anatomical segment $i$, let $\mathcal{V}_i$ denote the selected candidate surface vertices on the Nymeria template body mesh.
We compute a segment centroid $\mathbf{c}_i$ as the weighted average of the selected template vertices in $\mathcal{V}_i$.

To define a consistent in-surface direction, we set an anatomical axis $\mathbf{u}_i$ from $\mathbf{c}_i$ toward the centroid of its nearest available child segment in the body kinematic tree, or along the opposite direction from its nearest available parent when no child segment is available.
For each vertex $v\in\mathcal{V}_i$, we compute a surface normal $\mathbf{n}_{i,v}$ from the template mesh faces.
The normal defines a local tangent plane.
We choose the tangent direction $\mathbf{t}_{i,v}$ by projecting $\mathbf{u}_i$ onto this plane, set the binormal $\mathbf{b}_{i,v}=\mathbf{n}_{i,v}\times\mathbf{t}_{i,v}$, and form a right-handed surface-based sensor frame $\mathbf{R}^{\mathrm{surf}}_{i,v}=[\mathbf{t}_{i,v},\mathbf{b}_{i,v},\mathbf{n}_{i,v}]$:
\begin{equation}
\mathbf{t}_{i,v} =
\left(\mathbf{u}_i-(\mathbf{u}_i^\top\mathbf{n}_{i,v})\mathbf{n}_{i,v}\right)
/\left\|\mathbf{u}_i-(\mathbf{u}_i^\top\mathbf{n}_{i,v})\mathbf{n}_{i,v}\right\|_2.
\label{eq:surface_tangent}
\end{equation}
Let $\mathbf{p}_i(t)$, $\mathbf{R}_i(t)$, and $\mathbf{m}_v(t)$ denote the global segment position, global segment orientation, and posed mesh vertex position.
We estimate the local virtual sensor offset by $\mathbf{r}_{i,v}=\frac{1}{T}\sum_{t=1}^{T}\mathbf{R}_i(t)^\top(\mathbf{m}_v(t)-\mathbf{p}_i(t))$.
To account for mounting orientation variation during synthesis, we sample an in-plane rotation $\Delta\mathbf{R}_{i,v}$ around the surface normal and obtain the final local sensor frame $\widetilde{\mathbf{R}}^{\mathrm{surf}}_{i,v}=\Delta\mathbf{R}_{i,v}\mathbf{R}^{\mathrm{surf}}_{i,v}$.
The virtual IMU trajectory is then defined by its global position
$\mathbf{p}^{\mathrm{imu}}_{i,v}(t)=\mathbf{p}_i(t)+\mathbf{R}_i(t)\mathbf{r}_{i,v}$
and orientation
$\mathbf{R}^{\mathrm{imu}}_{i,v}(t)=\mathbf{R}_i(t)\widetilde{\mathbf{R}}^{\mathrm{surf}}_{i,v}$.
The accelerometer $\mathbf{a}_{i,v}(t)$ is obtained by transforming the second-order derivative of the virtual sensor position into the local sensor frame and removing gravity, while the gyroscope $\boldsymbol{\omega}_{i,v}(t)$ is computed from the temporal change of the virtual sensor orientation:
\begin{equation}
\begin{aligned}
\mathbf{a}_{i,v}(t)
&=
\mathbf{R}^{\mathrm{imu}}_{i,v}(t)^\top
\left(
d^2\mathbf{p}^{\mathrm{imu}}_{i,v}(t)/dt^2
-\mathbf{g}
\right)
+
\boldsymbol{\eta}^{a}_{i,v}(t), \\
\boldsymbol{\omega}_{i,v}(t)
&=
\mathbf{R}^{\mathrm{imu}}_{i,v}(t)^\top
\Omega\left(\mathbf{R}^{\mathrm{imu}}_{i,v}\right)(t)
+
\boldsymbol{\eta}^{\omega}_{i,v}(t).
\end{aligned}
\label{eq:sim_imu}
\end{equation}
Here $\mathbf{g}$ denotes gravity, $\Omega(\cdot)$ maps an orientation trajectory to angular velocity, and $\boldsymbol{\eta}^{a}_{i,v}(t)$ and $\boldsymbol{\eta}^{\omega}_{i,v}(t)$ denote accelerometer and gyroscope noise.
To reflect real device variability, we estimate two hardware-style noise priors from quiet windows of two real Nymeria IMU streams and randomly assign these priors to synthetic placements.
The final synthetic IMU candidate for placement $(i,v)$ is $\mathbf{x}_{i,v}(t)=[a^{x}_{i,v}(t),a^{y}_{i,v}(t),a^{z}_{i,v}(t),\omega^{x}_{i,v}(t),\omega^{y}_{i,v}(t),\omega^{z}_{i,v}(t)]$.
Collecting $\mathbf{x}_{i,v}$ over all selected vertices and anatomical segments yields a dense, geometry-aware distribution of wearable setups for pre-training. We evaluate the contribution of this geometry-aware simulation design in \autoref{tab:ablation}.

\subsection{Geometry-Aware Setup-Agnostic Pre-Training}
\label{sec:method_pretraining}
\begin{figure}[h]
% \vspace{-0.7em}
\centering
\includegraphics[width=\linewidth]{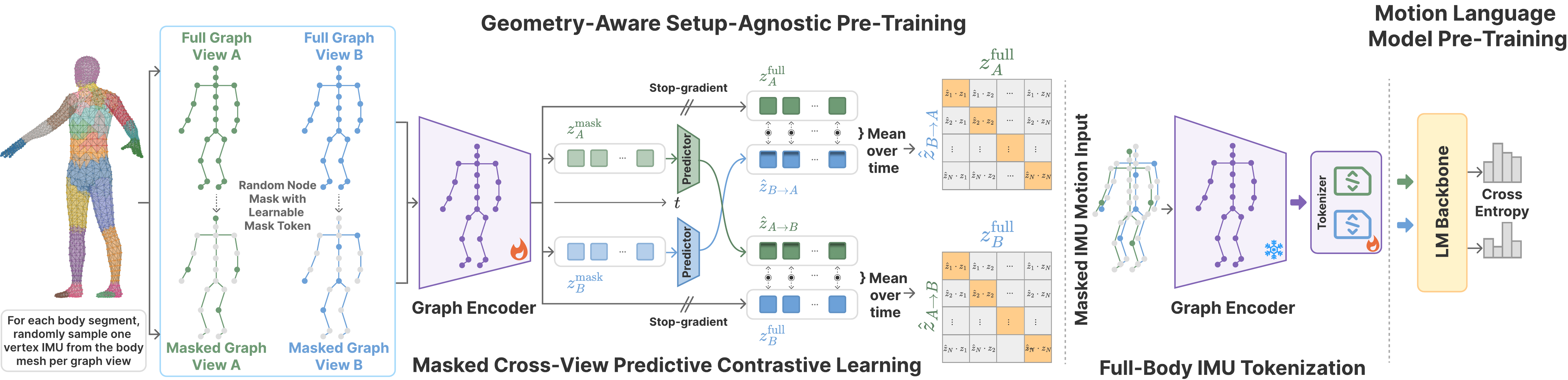}
\caption{Details of (1) Geometry-Aware Pre-Training, (2) Full-Body IMU Tokenization and (3) Motion Language Model Pre-Training.}
\label{fig:pretraining}
% \vspace{-0.7em}
\end{figure}

The dense simulation in \autoref{sec:method_simulation} provides multiple plausible IMU candidates for the same body motion, but downstream wearable inputs are sparse and setup-specific.
Even within the same body segment, different surface placements and sensor orientations can produce different IMU windows.
These setup variations are nevertheless organized by a fixed body topology: each synthetic IMU candidate is associated with one Nymeria anatomical segment, and segment motions are coupled through the body kinematic tree.
We use this structure to represent a full-body IMU observation as a spatio-temporal graph, where node $i$ stores the IMU window sampled for segment $i$ and edges follow the Nymeria kinematic tree.
Following spatial-temporal graph convolutional networks~\citep{yan2018spatial}, the graph encoder models temporal dynamics and cross-segment motion correlations while treating surface placement and mounting orientation as within-segment setup variation.
This requires a representation that remains stable across synthetic setup changes while retaining temporal motion structure beyond coarse activity labels.
As shown in \autoref{fig:pretraining}, we sample paired synthetic placement views from these candidates to train a setup-agnostic graph encoder %(\autoref{sec:method_mcvpcl}) 
and a full-body IMU tokenizer. %(\autoref{sec:method_tokenization}).

% \subsubsection{Masked Cross-View Predictive Contrastive Learning}
% \label{sec:method_mcvpcl}
\textbf{Masked Cross-View Predictive Contrastive Learning.} A natural choice is to contrast the two full graph views, following graph and skeleton contrastive representation learning~\citep{huang2023graph,lin2023actionlet, li20213d}.
However, full-view graph contrast can be satisfied by aligning complete body observations, so it does not teach the encoder to infer full-body motion from sparse wearable inputs.
Masked modeling methods address sparsity by reconstructing masked graph or motion tokens~\citep{yan2023skeletonmae, mao2023masked, hou2023graphmae2}, but masked prediction alone does not explicitly separate different motion instances.
Moreover, collapsing a motion window into a single clip-level embedding would remove the temporal structure needed by the tokenizer.
We therefore design a masked cross-view predictive contrastive objective that combines sparse-to-full recovery, robustness to within-segment placement and mounting orientation changes, contrastive discrimination, and time-preserving sequence latents.
Concretely, we predict the full-view latent of one synthetic setup from the masked observation of another setup, and contrast the prediction against other motion windows in the batch.
We further analyze the importance of this learning objective through ablation and embedding visualization in \autoref{tab:ablation} and \autoref{fig:umap}.

As illustrated in \autoref{fig:pretraining}, for each motion window we construct two full graph views $A$ and $B$ by independently sampling candidate placements $v_i^A,v_i^B\in\mathcal{V}_i$ for every segment $i$.
For each selected placement, we sample one local mounting rotation $\Delta\mathbf{R}^{A}_{i}$ or $\Delta\mathbf{R}^{B}_{i}$.
The rotation combines in-plane rotation around the surface normal with a small tilt around the local tangent axes to approximate imperfect surface attachment.
For an IMU candidate $\mathbf{x}_{i,v}(t)=[\mathbf{a}_{i,v}(t);\boldsymbol{\omega}_{i,v}(t)]$, rotation augmentation applies the same local rotation to acceleration and angular velocity, yielding paired full graph views $\mathbf{X}^{A},\mathbf{X}^{B}\in\mathbb{R}^{T\times N_{\mathrm{seg}}\times 6}$, where
$\mathbf{X}^{A}_{t,i,:}=[\Delta\mathbf{R}^{A}_{i}\mathbf{a}_{i,v_i^A}(t);\Delta\mathbf{R}^{A}_{i}\boldsymbol{\omega}_{i,v_i^A}(t)]$
and
$\mathbf{X}^{B}_{t,i,:}=[\Delta\mathbf{R}^{B}_{i}\mathbf{a}_{i,v_i^B}(t);\Delta\mathbf{R}^{B}_{i}\boldsymbol{\omega}_{i,v_i^B}(t)]$.
We then create masked graph views by randomly keeping between one and five visible segment nodes and replacing all other segment nodes with a learnable mask token, as shown in \autoref{fig:pretraining}.
Let $\mathcal{M}^{A}$ and $\mathcal{M}^{B}$ denote the visible segment sets.
The shared spatio-temporal graph encoder $\mathcal{G}_{\theta}$ produces node-level sequence features and averages over segment nodes to obtain time-preserving latents in $\mathbb{R}^{T'\times d}$, where $T'$ is the encoder output length and $d$ is the latent dimension.
We write the full-view latents as $\mathbf{z}^{\mathrm{full}}_{A}=\mathcal{G}_{\theta}(\mathbf{X}^{A})$ and $\mathbf{z}^{\mathrm{full}}_{B}=\mathcal{G}_{\theta}(\mathbf{X}^{B})$, and the masked-view latents as $\mathbf{z}^{\mathrm{mask}}_{A}=\mathcal{G}_{\theta}(\operatorname{Mask}(\mathbf{X}^{A},\mathcal{M}^{A}))$ and $\mathbf{z}^{\mathrm{mask}}_{B}=\mathcal{G}_{\theta}(\operatorname{Mask}(\mathbf{X}^{B},\mathcal{M}^{B}))$.
A temporal predictor $q_{\phi}$, implemented as a six-layer Transformer~\citep{vaswani2017attention}, predicts the opposite full-view latent: $\hat{\mathbf{z}}_{A\rightarrow B}=q_{\phi}(\mathbf{z}^{\mathrm{mask}}_{A})$ and $\hat{\mathbf{z}}_{B\rightarrow A}=q_{\phi}(\mathbf{z}^{\mathrm{mask}}_{B})$.

We train the encoder and predictor with a cross-view predictive InfoNCE loss, using mean cosine similarity over time for predicted and target sequence latents $\hat{\mathbf{z}},\mathbf{z}\in\mathbb{R}^{T'\times d}$, defined as $s(\hat{\mathbf{z}},\mathbf{z})=\frac{1}{T'}\sum_{\ell=1}^{T'}\frac{\hat{\mathbf{z}}_{\ell}^{\top}\mathbf{z}_{\ell}}{\|\hat{\mathbf{z}}_{\ell}\|_2\|\mathbf{z}_{\ell}\|_2}$.
For a minibatch of $N_{\mathrm{batch}}$ windows, the $A\rightarrow B$ loss is:
\begin{equation}
\mathcal{L}_{A\rightarrow B}
=
-\frac{1}{N_{\mathrm{batch}}}
\sum_{n=1}^{N_{\mathrm{batch}}}
\log
\frac{
\exp(s(\hat{\mathbf{z}}_{A\rightarrow B}^{(n)},\operatorname{sg}(\mathbf{z}_{B}^{\mathrm{full},(n)}))/\tau)
}{
\sum_{m=1}^{N_{\mathrm{batch}}}
\exp(s(\hat{\mathbf{z}}_{A\rightarrow B}^{(n)},\operatorname{sg}(\mathbf{z}_{B}^{\mathrm{full},(m)}))/\tau)
}.
\label{eq:mcvpcl_loss_ab}
\end{equation}
The $B\rightarrow A$ loss is defined symmetrically, and the final objective is
$\mathcal{L}_{\mathrm{MCVPCL}}
=
\mathcal{L}_{A\rightarrow B}
+
\mathcal{L}_{B\rightarrow A}$.
Here $\operatorname{sg}(\cdot)$ denotes stop-gradient and $\tau$ is temperature, $(n)$ and $(m)$ index windows in the minibatch.

% \subsubsection{Full-Body IMU Tokenization}
% \label{sec:method_tokenization}
\textbf{Full-Body IMU Tokenization.} Pre-training produces continuous, setup-stable sequence latents, while the motion-language model requires compact discrete inputs.
As motivated in \autoref{sec:introduction}, feeding raw IMU streams into an LLM is inefficient.
To connect wearable motion with language models, we train a full-body IMU tokenizer that discretizes the frozen graph encoder latent into compact IMU tokens, following recent motion-language tokenizers based on product quantization~\citep{hong2025egolm}.

As shown in \autoref{fig:pretraining}, with additional details in \autoref{fig:tokenization_pretraining_detail}, we freeze $\mathcal{G}_{\theta}$ and train a product-quantized VAE tokenizer on the latent obtained from masked wearable observations $\mathbf{z}^{\mathrm{mask}}=\mathcal{G}_{\theta}(\operatorname{Mask}(\mathbf{X},\mathcal{M}))\in\mathbb{R}^{T'\times d}$.
A projection maps each timestep latent $\mathbf{z}^{\mathrm{mask}}_{\ell}$ to a lower-dimensional projected latent $\bar{\mathbf{z}}_{\ell}\in\mathbb{R}^{\bar d}$.
Let $P$ denote the number of product codebooks and $K$ denote the number of entries in each codebook.
We evenly split $\bar{\mathbf{z}}_{\ell}$ into $P$ chunks $\{\bar{\mathbf{z}}_{\ell,j}\in\mathbb{R}^{\bar d/P}\}_{j=1}^{P}$, where $j$ indexes the product subspace.
The $j$-th codebook is $\mathcal{E}_{j}=\{\mathbf{e}_{j,k}\in\mathbb{R}^{\bar d/P}\}_{k=1}^{K}$, where $\mathbf{e}_{j,k}$ is the $k$-th code vector in that codebook.
Each chunk is quantized to its nearest code vector:
$\kappa_{\ell,j}=\arg\min_{k\in\{1,\ldots,K\}}\|\bar{\mathbf{z}}_{\ell,j}-\mathbf{e}_{j,k}\|_2^2$,
and the concatenated quantized latent
$\mathbf{q}_{\ell}=[\mathbf{e}_{1,\kappa_{\ell,1}};\ldots;\mathbf{e}_{P,\kappa_{\ell,P}}]$,
where $\kappa_{\ell,j}$ is the discrete code index for timestep $\ell$ and product subspace $j$, and $\mathbf{q}_{\ell}$ is the concatenated quantized latent.
A temporal convolutional decoder takes quantized sequence $\mathbf{q}=[\mathbf{q}_{1},\ldots,\mathbf{q}_{T'}]$ and reconstructs $\tilde{\mathbf{z}}=D(\mathbf{q})$.
We optimize the tokenizer with a reconstruction and commitment objective:
\begin{equation}
\mathcal{L}_{\mathrm{com}}
=
\sum_{j=1}^{P}
\frac{P}{T'\bar d}
\sum_{\ell=1}^{T'}
\left\|
\bar{\mathbf{z}}_{\ell,j}
-
\operatorname{sg}(\mathbf{e}_{j,\kappa_{\ell,j}})
\right\|_2^2,
\quad
\mathcal{L}_{\mathrm{tok}}
=
\operatorname{SmoothL1}(\tilde{\mathbf{z}},\mathbf{z}^{\mathrm{mask}})
+
\lambda_{\mathrm{com}}\mathcal{L}_{\mathrm{com}}.
\label{eq:tokenizer_loss}
\end{equation}    
We update codebooks with exponential moving averages and refresh dead codes to maintain codebook usage.
Finally, the IMU token sequence is formed by interleaving product-code indices over time,
$\mathbf{s}=[\kappa_{1,1},\ldots,\kappa_{1,P},\kappa_{2,1},\ldots,\kappa_{T',P}]$.
These discrete tokens preserve the temporal order of the setup-stable motion latent and serve as the IMU input tokens for motion-language alignment in \autoref{sec:method_motion_language}.
Since both $\mathcal{G}_{\theta}$ and the tokenizer operate temporally, variable-length IMU windows are handled by producing variable-length token sequences.

\subsection{Motion-Language Modeling}
\label{sec:method_motion_language}
The tokenizer in \autoref{sec:method_pretraining} converts sparse wearable observations into discrete IMU token sequences, but these new tokens are not yet meaningful to a pretrained LLM.
We therefore use motion language model pre-training to introduce the IMU vocabulary into the LLM and teach the model to understand wearable motion tokens.
Multi-task contrastive instruction tuning aligns IMU-token prompts with language descriptions and activity-label prompts for retrieval, captioning, and zero-shot recognition.

% \subsubsection{Motion Language Model Pre-Training}
% \label{sec:method_mlm_pretraining}
\textbf{Motion Language Model Pre-Training.} As shown in \autoref{fig:pretraining}, with details in \autoref{fig:tokenization_pretraining_detail}, motion language model pre-training adapts the LLM to the IMU token vocabulary.
We extend the LLM vocabulary with IMU tokens and one code token for each entry in each product codebook.
Rather than treating these tokens as unrelated new words, we use the learned tokenizer codebooks to give each IMU token a motion-aware embedding.
Specifically, for each IMU code token associated with a codebook vector $\mathbf{e}_{j,k}$, a projector $g_{\rho}$ maps $\mathbf{e}_{j,k}$ into the LLM embedding space, and the corresponding input embedding is replaced by $g_{\rho}(\mathbf{e}_{j,k})$ during model execution.
The same projected vectors are also used to initialize the corresponding rows in the LM head.
We then continue causal LM pre-training on the interleaved IMU token sequence $\mathbf{s}$ using next-token cross-entropy loss $\mathcal{L}_{\mathrm{CE}}=\operatorname{CE}(p(\cdot\mid s_{<r}),s_r)$.

\begin{figure}[h]
    % \vspace{-0.7em}
    \centering
    \includegraphics[width=\linewidth]{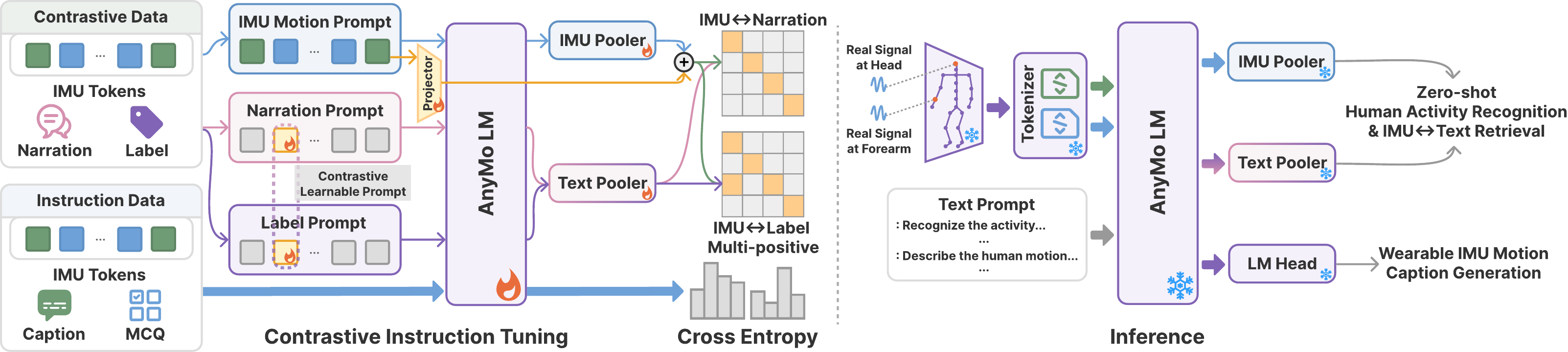}
    \caption{Details of Contrastive Instruction Tuning (left) and inference phases (right) of AnyMo.}
    \label{fig:cit_inference}
    % \vspace{-0.7em}
\end{figure}
% \subsubsection{Motion Language Multi-Task Contrastive Instruction Tuning}
% \label{sec:method_cit}
\textbf{Motion Language Multi-Task Contrastive Instruction Tuning.} Motion language model pre-training teaches the LLM to read IMU token sequences, but next-token prediction alone does not provide the discriminative motion-language alignment needed by retrieval and zero-shot recognition.
At the same time, captioning still requires the model to preserve its generative capability.
We therefore use multi-task contrastive instruction tuning to jointly support embedding-based and generation-based motion-language tasks~\citep{muennighoff2025generative,yu2025cafe}, as illustrated in \autoref{fig:cit_inference}.

The language supervision comes from Nymeria~\citep{ma2024nymeria} atomic-action annotations: we use the atomic-action text as motion narration and derive activity-label names from our semi-automatically curated and human-verified ground-truth action labels.
To increase linguistic diversity, each motion window is paired with five augmented narrations, as detailed in \autoref{sec:motion_language_training_details}.
For each training example, the IMU branch wraps the token sequence $\mathbf{s}$ in a prompt that also describes the visible wearable locations, while the language side contains two supervision branches built on the same text pooler: a label branch for activity-label names and a narration branch for ground-truth or augmented motion narrations.
The shared LLM processes these prompts. Considering the prompt sensitivity of LLM, the narration and label branches prepend a shared learnable soft prompt before LLM processing~\citep{liu2024gpt}.
The soft-prompt tokens are optimized through the contrastive objectives~\citep{tan2026bisecle} but are masked out during pooling, serving as contrastively learned context for language-side representations. We analyze this prompt sensitivity in \autoref{sec:prompt_sensitivity}.
In parallel, we use an instruction-generation branch for captioning and multiple-choice activity recognition.
This branch pairs the IMU token sequence $\mathbf{s}$ with natural-language task prompts, such as caption-generation or multiple-choice question instructions, and supervises the target response tokens through the LM head with next-token cross-entropy loss $\mathcal{L}_{\mathrm{CE}}$.
The contrastive embeddings are produced by modality-specific lightweight Transformer-style latent-attention poolers~\citep{lee2025nvembed} and projection heads.
Each pooler operates on the relevant IMU-token or text span while masking out unrelated prompt tokens.
The resulting IMU embedding lightly incorporates~\cite{he2016deep} averaged projected code-token embeddings to retain fine-grained motion semantics from the learned IMU codebook, yielding normalized embeddings $\mathbf{h}^{\mathrm{imu}}$ and $\mathbf{h}^{\mathrm{text}}$.
For a minibatch of $N_{\mathrm{batch}}$ paired IMU-narration examples, the symmetric narration-level IMU-text contrastive loss is:

\begin{equation}
\begin{aligned}
\mathcal{L}_{\mathrm{ITC}}
=
-\frac{1}{2N_{\mathrm{batch}}}
\sum_{n=1}^{N_{\mathrm{batch}}}
\Bigg[
&
\log
\frac{
\exp((\mathbf{h}^{\mathrm{imu}}_{n})^{\top}\mathbf{h}^{\mathrm{text}}_{n}/\tau_{\mathrm{ml}})
}{
\sum_{m=1}^{N_{\mathrm{batch}}}
\exp((\mathbf{h}^{\mathrm{imu}}_{n})^{\top}\mathbf{h}^{\mathrm{text}}_{m}/\tau_{\mathrm{ml}})
}
\\
&+
\log
\frac{
\exp((\mathbf{h}^{\mathrm{text}}_{n})^{\top}\mathbf{h}^{\mathrm{imu}}_{n}/\tau_{\mathrm{ml}})
}{
\sum_{m=1}^{N_{\mathrm{batch}}}
\exp((\mathbf{h}^{\mathrm{text}}_{n})^{\top}\mathbf{h}^{\mathrm{imu}}_{m}/\tau_{\mathrm{ml}})
}
\Bigg],
\end{aligned}
\label{eq:cit_itc}
\end{equation}
where $\tau_{\mathrm{ml}}$ is the motion-language contrastive temperature.
In addition to narration-level alignment, the label branch encodes activity-label names and applies a supervised label contrastive loss $\mathcal{L}_{\mathrm{label}}$, where all examples with the same normalized activity label are treated as positives.
The final objective is $\mathcal{L}_{\mathrm{CIT}}=\lambda_{\mathrm{CE}}\mathcal{L}_{\mathrm{CE}}+\lambda_{\mathrm{ITC}}\mathcal{L}_{\mathrm{ITC}}+\lambda_{\mathrm{label}}\mathcal{L}_{\mathrm{label}}$.
We further analyze the contributions of narration contrastive tuning, label contrastive tuning, and the multiple-choice question in the instruction branch in \autoref{tab:ablation}.
At inference time, a real sparse IMU window is converted into IMU tokens with its available wearable locations, and the tuned model is used either through the pooled embeddings for zero-shot recognition and IMU-text retrieval, or through the LM head for motion caption generation.

%% file: sec/4_experiments.tex
% !TEX root = ../neurips_2026.tex

\section{Experiments}
\label{sec:experiments}

\begin{table}[h]
% \vspace{-1em}
\caption{Zero-shot HAR comparison. The best is in \textbf{bold}, while the second-best is \underline{underlined}.}
\label{tab:zeroshot_har}
\centering
\scriptsize
\setlength{\tabcolsep}{3pt}

\renewcommand{\arraystretch}{1}
\begin{tabular}{c|c|*{6}{c}|*{5}{c}|*{3}{c}|c}
\toprule
\raisebox{1.2em}{\begin{tabular}{@{}c@{\hspace{0.35em}}|@{\hspace{0.35em}}c@{}}\rotatebox{75}{Method} & \rotatebox{75}{Dataset}\end{tabular}} & \rotatebox{75}{Metrics} 
& \rotatebox{75}{Opportunity}
& \rotatebox{75}{UCI-HAR}
& \rotatebox{75}{w-HAR}
& \rotatebox{75}{RealWorld}
& \rotatebox{75}{TNDA-HAR}
& \rotatebox{75}{EgoExo4D}
& \rotatebox{75}{OpenPack}
& \rotatebox{75}{PAMAP2}
& \rotatebox{75}{USC-HAD}
& \rotatebox{75}{WISDM}
& \rotatebox{75}{DSADS}
& \rotatebox{75}{UTD-MHAD}
& \rotatebox{75}{Ego4D}
& \rotatebox{75}{MMEA}
& \multicolumn{1}{|c}{\rotatebox{75}{Average}} \\
\midrule
\multicolumn{2}{c|}{Number of Classes}
& 4 & 6 & 7 & 8 & 8 & 8 & 10 & 12 & 12 & 18 & 19 & 27 & 31 & 32 & \\
\midrule
\multicolumn{2}{c|}{Level}
& \multicolumn{6}{c|}{Easy}
& \multicolumn{5}{c|}{Medium}
& \multicolumn{3}{c|}{Hard}
& \multicolumn{1}{|c}{Average} \\
\midrule

\multirow{3}{*}{ImageBind}
& Acc & \underline{59.3} & 14.0 & 8.2 & 17.4 & 22.6 & 8.2 & \underline{11.6} & 12.8 & 11.2 & 6.9 & 6.2 & 2.3 & 2.8 & 4.1 & 13.4 \\
& F1  & 39.6 & 6.7 & 6.1 & 10.4 & 18.6 & 8.5 & 8.3 & 7.7 & 6.9 & 4.4 & 3.6 & 0.8 & 0.8 & 1.3 & 8.8 \\
& R@2 & \underline{83.5} & 20.2 & 52.5 & 27.4 & 35.3 & 12.5 & 20.7 & 15.5 & 23.5 & 13.5 & 8.2 & 4.7 & 6.4 & 7.6 & 23.7 \\
\midrule

\multirow{3}{*}{IMU2CLIP}
& Acc & 47.3 & 30.1 & 36.1 & 27.3 & 26.8 & 10.9 & 9.5 & 14.0 & 14.8 & 12.4 & 11.5 & 3.7 & 1.8 & 5.3 & 18.0 \\
& F1  & 37.6 & \underline{27.3} & 20.9 & 20.5 & 22.3 & 5.9 & 4.5 & 11.3 & 11.6 & 8.3 & 8.3 & 0.4 & 0.4 & \underline{3.1} & 13.0 \\
& R@2 & 69.0 & 55.1 & 60.7 & 43.9 & 46.2 & 18.5 & 19.5 & 26.6 & 29.1 & 20.4 & 17.6 & 7.9 & 3.1 & \underline{10.8} & 30.6 \\
\midrule

\multirow{3}{*}{IMUGPT}
& Acc & 10.1 & 1.1 & \textbf{67.2} & 16.9 & 14.3 & 12.5 & 11.4 & 8.9 & 6.0 & 8.3 & 7.5 & 3.7 & \underline{4.6} & 3.4 & 12.6 \\
& F1  & 10.4 & 0.3 & 38.8 & 4.0 & 6.1 & 7.6 & \underline{9.1} & 1.5 & 6.9 & 6.6 & 2.0 & 0.3 & 1.9 & 1.8 & 7.0 \\
& R@2 & 33.7 & 18.2 & \underline{67.2} & 33.7 & 28.5 & 27.4 & \underline{22.9} & 19.3 & 31.8 & 13.9 & 14.6 & 8.8 & \underline{9.3} & 6.7 & 24.0 \\
\midrule

\multirow{3}{*}{HARGPT}
& Acc & 28.8 & 15.0 & 4.9 & 12.7 & 13.7 & 16.1 & 10.2 & 11.1 & 9.5 & 5.5 & 5.8 & 3.3 & 3.6 & 2.4 & 10.2 \\
& F1  & 17.3 & 12.7 & 3.1 & 5.3 & 5.4 & 12.0 & 5.5 & 2.1 & 3.6 & 3.5 & 3.4 & 1.5 & 1.1 & 0.9 & 5.5 \\
& R@2 & 47.0 & 31.4 & 11.5 & 31.7 & 25.2 & 32.0 & 22.7 & 23.0 & 17.6 & 11.8 & 12.1 & 9.3 & 7.8 & 6.3 & 20.7 \\
\midrule

\multirow{3}{*}{UniMTS}
& Acc & 45.9 & \underline{35.2} & \underline{59.0} & \underline{43.6} & \underline{59.1} & 23.1 & 11.5 & \underline{47.2} & \underline{30.5} & \textbf{27.8} & \underline{31.5} & \textbf{22.8} & 3.7 & \underline{6.1} & \underline{31.9} \\
& F1  & \underline{42.2} & 22.0 & \textbf{42.9} & \underline{36.7} & \textbf{53.7} & \underline{18.4} & 7.5 & \textbf{43.6} & \textbf{27.8} & \textbf{25.5} & \underline{23.7} & \textbf{18.5} & \underline{4.3} & 2.8 & \underline{26.4} \\
& R@2 & 80.0 & 53.1 & 60.7 & \underline{64.0} & \underline{77.5} & \underline{47.0} & 21.9 & \underline{63.2} & 45.4 & \textbf{47.1} & \underline{46.0} & \textbf{32.6} & 6.9 & 10.7 & \underline{46.9} \\
\midrule

\multirow{3}{*}{NormWear}
& Acc & 26.0 & 3.7 & 3.3 & 16.8 & 12.2 & 10.5 & 9.8 & 7.9 & 8.7 & 4.4 & 0.7 & 3.7 & 2.0 & 2.7 & 8.0 \\
& F1  & 10.3 & 1.6 & 1.3 & 3.8 & 2.8 & 3.1 & 3.1 & 1.6 & 1.4 & 0.9 & 0.1 & 0.3 & 0.2 & 0.3 & 2.2 \\
& R@2 & 66.1 & 29.6 & 3.3 & 20.4 & 19.1 & 20.2 & 16.5 & 10.5 & 12.2 & 12.4 & 2.6 & 7.4 & 5.8 & 6.0 & 16.6 \\
\midrule

\multirow{3}{*}{$\text{Gemma 4 26B}_{\text{ Text}}$}
& Acc & 35.9 & 29.4 & 29.5 & 30.8 & 28.2 & 18.0 & 9.6 & 13.3 & 29.7 & 11.6 & 10.7 & 4.7 & 2.4 & 5.6 & 18.5 \\
& F1  & 23.5 & 19.9 & 12.7 & 18.8 & 19.3 & 11.8 & 7.6 & 7.4 & 12.4 & 8.0 & 7.6 & 2.0 & 1.2 & 1.9 & 11.0 \\
& R@2 & 68.4 & \underline{60.5} & 31.1 & 58.9 & 53.1 & 40.7 & 19.4 & 31.1 & 42.7 & 20.8 & 22.0 & 10.2 & 6.7 & 9.4 & 33.9 \\
\midrule

\multirow{3}{*}{$\text{Gemma 4 26B}_{\text{ Plot}}$}
& Acc & 39.8 & 29.8 & 27.9 & 31.2 & 27.8 & \underline{24.8} & 7.7 & 19.4 & \textbf{32.9} & 9.4 & 11.5 & 5.6 & 3.5 & 4.4 & 19.7 \\
& F1  & 33.0 & 21.6 & 10.6 & 23.6 & 15.1 & 11.9 & 4.5 & 10.8 & 14.0 & 6.1 & 7.5 & 1.2 & 1.3 & 2.0 & 11.7 \\
& R@2 & 74.4 & 59.6 & 32.8 & 58.1 & 50.1 & 24.8 & 18.2 & 35.4 & \underline{46.8} & 16.9 & 21.5 & 10.7 & 7.9 & 7.9 & 33.2 \\
\midrule

\multirow{3}{*}{\textbf{AnyMo}}
& Acc & \textbf{59.4} & \textbf{56.5} & 57.4 & \textbf{48.4} & \textbf{59.4} & \textbf{30.2} & \textbf{13.1} & \textbf{52.6} & 27.7 & \underline{25.4} & \textbf{36.3} & \underline{16.3} & \textbf{8.6} & \textbf{8.1} & \textbf{35.7}{\textcolor{teal}{\textbf{(+11.7\%)}}} \\
& F1  & \textbf{58.8} & \textbf{51.6} & \underline{42.2} & \textbf{37.2} & \underline{53.1} & \textbf{24.1} & \textbf{11.6} & \underline{41.5} & \underline{22.6} & \underline{18.6} & \textbf{29.5} & \underline{11.3} & \textbf{6.3} & \textbf{4.0} & \textbf{29.5}{\textcolor{teal}{\textbf{(+11.6\%)}}} \\
& R@2 & \textbf{83.5} & \textbf{89.5} & \textbf{98.4} & \textbf{77.6} & \textbf{87.9} & \textbf{51.6} & \textbf{28.0} & \textbf{78.2} & \textbf{64.0} & \underline{41.1} & \textbf{53.0} & \underline{24.2} & \textbf{13.7} & \textbf{13.9} & \textbf{57.5}{\textcolor{teal}{\textbf{(+22.6\%)}}} \\

\bottomrule
\end{tabular}
% \vspace{-1em}
\end{table}

% \subsection{Experiment Setups}
We train AnyMo using Nymeria~\citep{ma2024nymeria}, which provides synchronized body mesh/skeleton motion, atomic-action text annotations, and real IMU streams from the head and two wrists.
The mesh and skeleton motion are used to simulate dense geometry-aware IMU candidates for pre-training, while real Nymeria IMU streams are used only to estimate device-noise priors and for held-out sim-to-real evaluation.
No downstream benchmark dataset is used to train AnyMo.
We use Qwen2.5-0.5B as the LLM backbone of AnyMo~\citep{qwen2025qwen25technicalreport}.
We evaluate along three complementary axes: zero-shot activity recognition across 14 unseen wearable datasets, bidirectional IMU-text retrieval, and wearable IMU motion caption generation.
At inference time, each sparse IMU window is converted into IMU tokens, and AnyMo uses only these tokens, the visible wearable-location context, and task-specific text prompts.
For recognition, the 14 completely unseen datasets~\citep{roggen2010collecting, anguita2013public, bhat2020w, sztyler2016body, yan2021tnda, grauman2024ego, yoshimura2024openpack, reiss2012introducing, zhang2012usc, weiss2019wisdm, altun2010comparative, chen2015utd, grauman2022ego4d, xu2024towards} cover diverse body locations, devices, sampling protocols, and activity vocabularies, including four large-scale in-the-wild datasets. We group them by label space size into easy (<10 classes), medium (10--20 classes), and hard (>20 classes) settings.
For retrieval and captioning, we evaluate sim-to-real transfer on held-out Nymeria subjects and OOD zero-shot transfer to EgoExo4D.
We compare against sensor-language multimodal methods~\citep{girdhar2023imagebind, moon2023imu2clip, zhang2024unimts}, synthetic IMU pre-training methods~\citep{leng2023generating,zhang2024unimts}, wearable foundation model~\citep{luo2026toward}, and multimodal LLM baselines~\citep{ji2024hargpt, openai2025gpt54thinking, googledeepmind2026gemma4modelcard}.
We report Accuracy, macro-F1, and Recall@2 for recognition; Recall@K and MRR for retrieval; and BLEU, ROUGE-L, METEOR, and BERT-F1 for captioning.
Full dataset statistics, preprocessing, prompt templates, and baseline details are provided in \autoref{sec:experiment_details}, \autoref{sec:downstream_evaluation_dataset}, and \autoref{sec:baseline_details}. Separately, \autoref{sec:anymo_bench} describes AnyMo Bench, an additional in-the-wild HAR benchmark built from our curated Nymeria activity labels for fine-grained unseen-subject and cross-device recognition.

% \subsection{Zero-shot Human Activity Recognition}

\textbf{Zero-shot Human Activity Recognition.}  \autoref{tab:zeroshot_har} summarizes zero-shot activity recognition on 14 completely unseen wearable datasets.
AnyMo achieves the best average performance across all three metrics, with 35.7 Accuracy, 29.5 macro-F1, and 57.5 Recall@2, improving over the strongest average baseline by 11.7\%, 11.6\%, and 22.6\%, respectively.
The gains hold across controlled HAR benchmarks and in the wild datasets, suggesting that AnyMo learns transferable motion-language representations.
ImageBind and IMU2CLIP provide sensor-language multimodal alignment and perform competitively on a few low-class datasets, but their performance drops as sensing setups and label spaces become more diverse.
IMUGPT and UniMTS benefit from synthetic IMU motion pre-training, with UniMTS serving as the strongest prior baseline across the benchmark. AnyMo further improves over them through geometry-aware surface simulation and pre-training, and motion-language alignment.
HARGPT and the Gemma 4 26B prompting baselines test whether substantially larger LLM priors alone can support zero-shot HAR: HARGPT uses role-play and step-by-step prompts over raw IMU readings, while Gemma 4 26B uses numerical IMU input or plots of IMU readings for multimodal understanding.
The Gemma 4 26B prompting baselines occasionally perform well on individual datasets, but remain substantially below AnyMo on average, indicating that direct language or vision-language prompting, even at larger scale, is not sufficient for robust cross-setup wearable motion recognition.
NormWear, despite pre-training on heterogeneous physiological and inertial signals including IMU, performs weakly in this benchmark. Suggesting that broad wearable-signal pre-training alone does not directly transfer to language-grounded open-vocabulary HAR across diverse unseen datasets and body placements.

% \subsection{Cross-Modal Retrieval}

\begin{table}[h]
% \vspace{-0.7em}
\caption{Unseen and zero-shot cross-modal retrieval performance on Nymeria held-out set and EgoExo4D datasets. ``--'' denotes settings infeasible for LLM baselines due to context-length limits.}
\label{tab:unseen_cross_modal_retrieval}
\centering
\scriptsize
\setlength{\tabcolsep}{2.2pt}
\renewcommand{\arraystretch}{1}
\resizebox{\linewidth}{!}{%
\begin{tabular}{c|l|*{4}{c}|*{4}{c}|*{4}{c}|*{4}{c}}
\toprule
\multicolumn{2}{c|}{Dataset}
& \multicolumn{8}{c|}{Nymeria Held-out}
& \multicolumn{8}{c}{EgoExo4D Zero-shot} \\
\cmidrule(lr){1-2}\cmidrule(lr){3-10}\cmidrule(lr){11-18}
\multicolumn{1}{c|}{\multirow{2}{*}{Task}}
& \multicolumn{1}{c|}{\multirow{2}{*}{Method}}
& \multicolumn{4}{c|}{100 Samples}
& \multicolumn{4}{c|}{All Samples}
& \multicolumn{4}{c|}{100 Samples}
& \multicolumn{4}{c}{All Samples} \\
\cmidrule(lr){3-6}\cmidrule(lr){7-10}\cmidrule(lr){11-14}\cmidrule(lr){15-18}
& & R@1 & R@5 & R@10 & MRR
& R@1 & R@5 & R@10 & MRR
& R@1 & R@5 & R@10 & MRR
& R@1 & R@5 & R@10 & MRR \\
\midrule
\multirow{6}{*}{\shortstack{IMU\\$\downarrow$\\Text}}
& ImageBind & 0.0 & 6.0 & 14.0 & 5.0 & 0.1 & 0.2 & 0.3 & 0.3 & 1.0 & 5.0 & 8.0 & 4.6 & 0.1 & 0.2 & 0.3 & 0.3 \\
& IMU2CLIP & 1.0 & 6.0 & 12.0 & 5.5 & 0.0 & 0.1 & 0.3 & 0.3 & \underline{2.0} & \textbf{10.0} & \underline{23.0} & \underline{8.2} & 0.0 & 0.3 & 0.5 & 0.4 \\
& UniMTS & \underline{4.0} & \underline{12.0} & \underline{23.0} & \underline{10.0} & \underline{0.2} & \underline{0.9} & \underline{1.6} & \underline{0.9} & 1.0 & 9.0 & 16.0 & 6.3 & \underline{0.1} & \underline{0.6} & \underline{1.3} & \underline{0.7} \\
& GPT-5.4 Mini & 1.0 & 7.0 & 11.0 & 4.4 & -- & -- & -- & -- & 1.0 & 6.0 & 10.0 & 3.7 & -- & -- & -- & -- \\
& Gemma 4 26B & 2.0 & 9.0 & 16.0 & 6.1 & -- & -- & -- & -- & 2.0 & 4.0 & 12.0 & 4.6 & -- & -- & -- & -- \\
& \textbf{AnyMo} & \textbf{22.0} & \textbf{64.0} & \textbf{78.0} & \textbf{40.1} & \textbf{2.7} & \textbf{8.8} & \textbf{14.9} & \textbf{7.0} & \textbf{2.0} & \underline{9.0} & \textbf{27.0} & \textbf{9.5} & \textbf{0.2} & \textbf{0.7} & \textbf{1.4} & \textbf{0.8} \\
\midrule
\multirow{6}{*}{\shortstack{Text\\$\downarrow$\\IMU}}
& ImageBind & \underline{1.0} & \underline{8.0} & \underline{14.0} & \underline{6.7} & \underline{0.1} & \underline{0.2} & 0.3 & 0.3 & \underline{2.0} & 3.0 & 7.0 & 5.1 & 0.0 & 0.0 & 0.2 & 0.2 \\
& IMU2CLIP & 0.0 & 6.0 & 14.0 & 5.0 & 0.1 & 0.2 & 0.3 & \underline{0.3} & 1.0 & \underline{9.0} & \underline{17.0} & \underline{7.7} & \textbf{0.1} & \textbf{0.3} & \underline{0.5} & \underline{0.4} \\
& UniMTS & 1.0 & 6.0 & 12.0 & 5.5 & 0.1 & 0.2 & \underline{0.4} & 0.3 & 1.0 & 5.0 & 10.0 & 5.3 & 0.0 & 0.1 & 0.3 & 0.2 \\
& GPT-5.4 Mini & -- & -- & -- & -- & -- & -- & -- & -- & -- & -- & -- & -- & -- & -- & -- & -- \\
& Gemma 4 26B & -- & -- & -- & -- & -- & -- & -- & -- & -- & -- & -- & -- & -- & -- & -- & -- \\
& \textbf{AnyMo} & \textbf{21.0} & \textbf{57.0} & \textbf{80.0} & \textbf{38.5} & \textbf{2.5} & \textbf{8.5} & \textbf{14.0} & \textbf{6.6} & \textbf{3.0} & \textbf{10.0} & \textbf{23.0} & \textbf{9.9} & \underline{0.0} & \underline{0.3} & \textbf{0.6} & \textbf{0.4} \\
\bottomrule
\end{tabular}
}
% \vspace{-0.7em}
\end{table}

\textbf{Cross-Modal Retrieval.} \autoref{tab:unseen_cross_modal_retrieval} evaluates bidirectional retrieval between sparse IMU windows and motion narrations.
The Nymeria held-out split is designed as a synthetic-to-real transfer test: AnyMo is trained without the held-out subjects using synthetic IMU candidates generated from body mesh/skeleton motion, while all methods are evaluated on the same real IMU streams from held-out subjects.
ImageBind and IMU2CLIP reflect large-scale real sensor-language pre-training, UniMTS reflects an alternative synthetic IMU pre-training strategy, and GPT-5.4 Mini/Gemma 4 26B test whether much larger state-of-the-art LLMs can solve retrieval through prompting.
AnyMo substantially outperforms all baselines on Nymeria held-out, improving 100-sample IMU$\rightarrow$Text MRR from 10.0 to 40.1 and Text$\rightarrow$IMU MRR from 6.7 to 38.5, while remaining clearly ahead in the harder all-sample setting.
EgoExo4D further evaluates OOD zero-shot transfer. Although all methods degrade, AnyMo achieves the best or second-best performance on most metrics.

\begin{figure}[h]
\centering
\includegraphics[width=\linewidth]{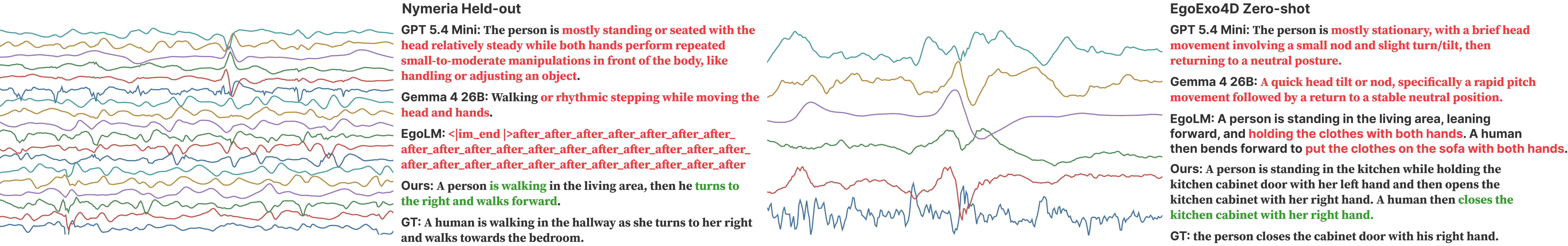}
\caption{Qualitative Results of Wearable IMU Motion Caption Generation. We use green to highlight correct parts and red for mistakes.}
\label{fig:qualitative_caption}
\end{figure}

% \textbf{Wearable IMU Motion Caption Generation.} 
\textbf{Wearable IMU Motion Captioning.} \autoref{tab:wearable_imu_caption_generation} evaluates AnyMo's generative motion-language capability.
Under the same Nymeria held-out and EgoExo4D zero-shot settings, AnyMo outperforms GPT-5.4 Mini, Gemma 4 26B and EgoLM across captioning metrics.
On Nymeria held-out, AnyMo substantially improves ROUGE-L from the best baseline score of 19.0 to 31.4 and BERT-F1 from 57.3 to 69.8.
These gains show that the learned IMU tokens retain motion semantics that can be decoded into natural-language descriptions.
The advantage persists under EgoExo4D zero-shot transfer, where AnyMo achieves 20.7 BLEU-1, 21.0 METEOR, and 67.1 BERT-F1 despite OOD motion distributions.
These results support AnyMo's captioning effectiveness relative to both direct LLM prompting and a trained motion-language baseline.
\autoref{fig:qualitative_caption} provides qualitative examples.
The prompting baselines often describe generic posture changes or local signal fluctuations, such as standing, head motion, or hand movement, while missing the full activity.
In contrast, AnyMo produces captions that better preserve the action-level semantics, including walking direction changes in Nymeria and cabinet-closing interactions in EgoExo4D.
These results suggest that the IMU tokenizer and motion-language instruction tuning enable open-ended motion description from wearable signals.

\begin{table}[h]
% \vspace{-0.7em}
\caption{Unseen and zero-shot IMU motion caption generation results on Nymeria and EgoExo4D.}
\label{tab:wearable_imu_caption_generation}
\centering
\scriptsize
\setlength{\tabcolsep}{3.2pt}
\renewcommand{\arraystretch}{1}
\begin{tabular}{l|*{5}{c}|*{5}{c}}
\toprule
\multicolumn{1}{c|}{Dataset}
& \multicolumn{5}{c|}{Nymeria Held-out}
& \multicolumn{5}{c}{EgoExo4D Zero-shot} \\
\cmidrule(lr){1-1}\cmidrule(lr){2-6}\cmidrule(lr){7-11}
\multicolumn{1}{c|}{Method}
& BLEU-1 & BLEU-4 & ROUGE-L & METEOR & BERT-F1
& BLEU-1 & BLEU-4 & ROUGE-L & METEOR & BERT-F1 \\
\midrule
GPT-5.4 Mini & 19.2 & 0.3 & 15.7 & 17.2 & 57.3 & 12.6 & 0.0 & 15.5 & 16.8 & 56.5 \\
Gemma 4 26B & 16.2 & 0.0 & 13.6 & 14.2 & 56.5 & 3.5 & 0.0 & 4.6 & 7.1 & 55.1 \\
EgoLM & 16.2 & 3.3 & 19.0 & 16.8 & 53.5 & 19.0 & 0.1 & 18.1 & 20.1 & 62.4 \\
\textbf{AnyMo} & \textbf{26.9} & \textbf{7.0} & \textbf{31.4} & \textbf{28.6} & \textbf{69.8} & \textbf{20.7} & \textbf{0.4} & \textbf{19.7} & \textbf{21.0} & \textbf{67.1} \\
\bottomrule
\end{tabular}
% \caption{Unseen/zero-shot IMU motion caption generation results on Nymeria and EgoExo4D.}
% \caption{Unseen and zero-shot IMU motion captioning results on Nymeria and EgoExo4D.}
% \caption{Unseen and zero-shot wearable IMU motion caption generation performance on Nymeria held-out set and EgoExo4D zero-shot datasets.}
% \vspace{-1em}
\end{table}

\begin{figure}[h]
    \centering
    \includegraphics[width=\linewidth]{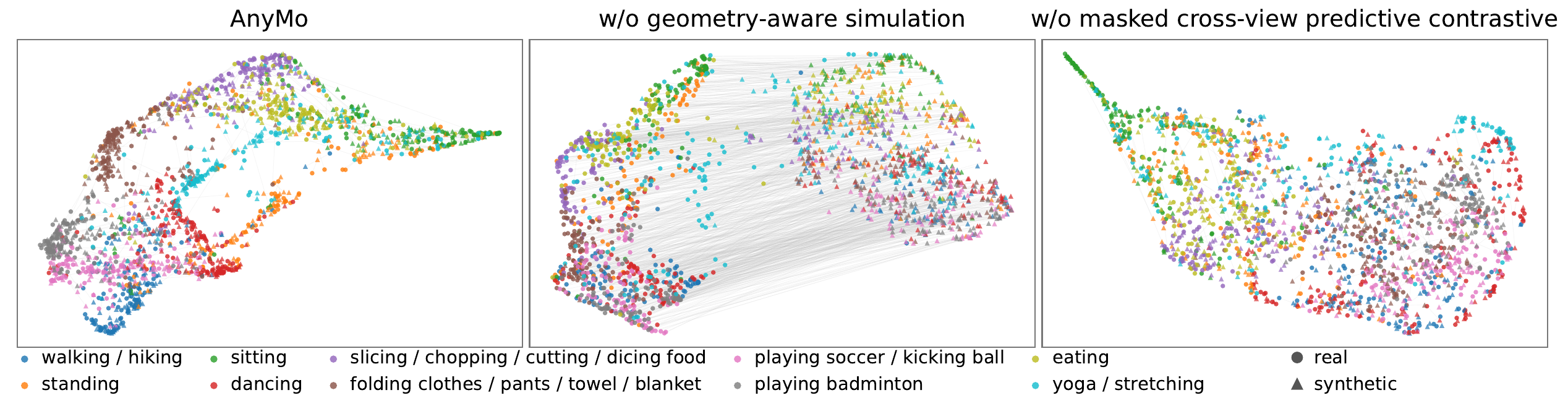}
    \caption{UMAP visualization of paired real and synthetic IMU embeddings for ten activity categories.}
    \label{fig:umap}
\end{figure}

% \subsection{Ablation Study}
\textbf{Ablation.} \autoref{tab:ablation} ablates each component (see \autoref{sec:ablation_implementation_details} for detailed implementation).
% \textbf{Ablation Study.} \autoref{tab:ablation} studies the contribution of each component.
% Implementation details for each ablation variant are provided in \autoref{sec:experiment_details}.
Removing geometry-aware simulation causes the largest degradation, reducing performance from 35.7/29.5/57.5 to 18.5/10.3/30.0 in Acc/F1/R@2.
Replacing the masked cross-view predictive contrastive objective also severely hurts performance.
These results indicate that synthetic IMU generation alone is insufficient: the model needs both realistic surface-aware setup variation and a sparse-to-full pre-training objective to transfer from synthetic full-body candidates to real sparse wearable inputs.
\autoref{fig:umap} further visualizes this effect.
For 10 activity categories, we randomly sample 100 real and 100 synthetic windows per activity and plot their embeddings with UMAP~\citep{mcinnes2018umap}.
\begin{wraptable}{r}{0.45\textwidth}
    \vspace{-0.5em}
    \caption{Ablation study.}
    \label{tab:ablation}
    \vspace{0.5em}
    \centering
    \scriptsize
    \setlength{\tabcolsep}{3.2pt}
    \renewcommand{\arraystretch}{1}
    \begin{tabular}{lccc}
    \toprule
    Variant & Acc & F1 & R@2 \\
    \midrule
    \textbf{AnyMo} & \textbf{35.7} & \textbf{29.5} & \textbf{57.5} \\
    \rowcolor{gray!12}
    \multicolumn{4}{l}{\emph{Simulation and encoder pretraining}} \\
    w/o geometry-aware simulation & 18.5 & 10.3 & 30.0 \\
    w/o masked cross-view pred. contrastive & 26.7 & 18.8 & 45.1 \\
    \rowcolor{gray!12}
    \multicolumn{4}{l}{\emph{Motion-language contrastive}} \\
    w/o label contrastive & 33.2 & \underline{26.7} & 48.3 \\
    w/o narration contrastive & 31.0 & 22.2 & 46.8 \\
    w/o all contrastive & 24.2 & 15.4 & 40.7 \\
    \rowcolor{gray!12}
    \multicolumn{4}{l}{\emph{LM instruction tuning}} \\
    w/o MCQ instruction tuning & \underline{33.4} & 26.2 & \underline{53.3} \\
    \bottomrule
    \end{tabular}
    % \vspace{-1.5em}
\end{wraptable}
AnyMo forms more discriminative activity clusters while aligning real and synthetic samples of the same activity.
Without geometry-aware simulation, real and synthetic embeddings separate by domain, while removing masked cross-view predictive contrastive learning reduces cluster coherence, highlighting their complementary roles in sim-to-real alignment and activity discrimination.
Controlled Nymeria experiments and cross-dataset evaluation across 17 real sensor locations quantitatively support robustness to setup variation and real-synthetic motion correspondence (see \autoref{sec:setup_transfer_alignment}).
The motion-language losses are also important. 
Removing label or narration contrastive tuning degrades Acc, F1, and R@2, showing that both label-level supervision and free-form narration alignment support zero-shot recognition.
Removing all contrastive tuning further degrades performance, confirming that next-token instruction tuning alone is insufficient for discriminative open-vocabulary recognition.
Finally, removing MCQ instruction tuning reduces Acc, F1, and R@2, indicating that MCQ supervision supports both activity discrimination and candidate ranking.

%% file: sec/5_conclusion.tex
% !TEX root = ../neurips_2026.tex

\section{Conclusion}
\label{sec:conclusion}

We presented AnyMo, a geometry-aware setup-agnostic framework for wearable IMU motion understanding under variable sensing setups. AnyMo treats wearable setup variation as structured body-surface variation, synthesizes dense physics-grounded IMU candidates from human mesh motion, learns setup-agnostic full-body motion representations, and connects them to language models through compact full-body IMU tokens.
Across zero-shot activity recognition, bidirectional IMU-text retrieval, and wearable motion caption generation, AnyMo shows consistent gains on unseen datasets, held-out real Nymeria IMU streams, and out-of-domain EgoExo4D transfer.
These results suggest that combining geometry-aware wearable simulation with motion-language modeling is a promising path toward generalist motion understanding from sparse wearable sensors.

%% file: sec/2_related_works.tex
% !TEX root = ../neurips_2026.tex
\section{Related Works}
\label{sec:related_work}

\textbf{Wearable Motion Representations and Setup Transfer.}
Recent wearable motion models improve generalization through joint-centric synthetic pretraining~\citep{zhang2024unimts, leng2024imugpt2, leng2023generating, miao2026wonderwall, wei2025one}, large-scale self-supervision or cross dataset adaptation~\citep{xu2025relcon, hong2024crosshar}, and tokenization~\citep{zhang2025mopformer}. 
Other works study setup variation via cross-location transfer~\citep{fortes2022learning}, body-surface placement ranking with location-specific classifiers~\citep{ray2025w2w}, or flexible placement conditioned on sensor coordinates at inference~\citep{zhou2025imucoco}; however, they remain recognition- or pose-centered rather than learning setup-robust full-body representations from sparse IMU for recognition and motion-language understanding without requiring within-segment coordinates.
\textbf{Sensor-Language and Cross-Modal Grounding.}
Sensor-language and multimodal methods connect IMU or human-sensing signals to language, LLM reasoning, shared embedding spaces, or cross-modal supervision~\citep{zhang2025sensorlm, luo2026toward, li2025sensorllm, ji2024hargpt, asif2025llasa, wang2024ubiphysio, su2025imuzero, moon2023imu2clip, girdhar2023imagebind, zhou2026holollm, leng2023generating, leng2024imugpt2, chen2025comodo, tong2022zero, li2025zara, hong2025egolm, wang2025ego4o, jiang2023motiongpt, nguyen2026mobind, kinfu2025motionbind}.
Embedding and instruction-tuning methods~\citep{yu2025cafe,lee2025nvembed, muennighoff2025generative} motivate joint retrieval and generation training, but these works do not directly address sparse, setup-dependent wearable IMU through geometry-aware full-body tokenization.
\textbf{Self-Supervised Representation Learning.}
Skeleton and graph self-supervised methods, physiological-signal and audio pretraining~\citep{yan2023skeletonmae, mao2023masked, huang2023graph, lin2023actionlet, hou2023graphmae2, li20213d, li2026glucofm, li2026audiofm}, motivate our use of masked motion modeling, graph encoders, and cross-view consistency, but these works do not jointly address sparse-to-full inference, within-segment placement robustness, and activity discrimination while preserving temporal structure for wearable IMU tokenization.

\textbf{Comparison Criteria.}
In \autoref{tab:conceptual_comparison}, partial support distinguishes
related capabilities from the full scope of each criterion.
For placement transfer, UniMTS and Wonderwall incorporate orientation
augmentation but do not explicitly model within-segment surface displacement.
CrossHAR supports cross-dataset transfer within a shared activity-label
space, rather than zero-shot recognition of unseen activity classes. For open-source availability, SensorLM releases caption-construction utilities and reference code for a SigLIP variant, but the public release does not include the full implementation or pretrained weights of the contrastive-generative
model evaluated in the paper. IMUGPT and IMUGPT 2.0 require target activity classes to generate textual
descriptions, synthesize motion and IMU data, and train a target-class
classifier before evaluation, rather than directly recognizing new
classes without retraining.
% \textbf{Self-Supervised Skeleton and Graph Motion Pretraining.}
% Skeleton and graph self-supervised methods~\citep{yan2023skeletonmae, mao2023masked, huang2023graph, lin2023actionlet, hou2023graphmae2, li20213d} motivate our use of masked motion modeling, graph encoders, and cross-view consistency, but these works focus on pose or skeleton observations rather than sparse-to-full motion inference across variable wearable setups.
%A detailed discussion of related work is provided in Appendix~\ref{sec:detail_related_works}.

%% file: ref.bib
@article{rosenhahn2008human,
  title={Human motion},
  author={Rosenhahn, Bodo and Klette, Reinhard and Metaxas, Dimitris},
  journal={Understanding, Modeling, Capture},
  year={2008},
  publisher={Springer}
}

@article{aggarwal2011human,
author = {Aggarwal, J.K. and Ryoo, M.S.},
title = {Human activity analysis: A review},
year = {2011},
issue_date = {April 2011},
publisher = {Association for Computing Machinery},
address = {New York, NY, USA},
volume = {43},
number = {3},
issn = {0360-0300},
doi = {10.1145/1922649.1922653},
journal = {ACM Comput. Surv.},
month = apr,
articleno = {16},
numpages = {43}
}

@inproceedings{schilit1994context,
  title={Context-aware computing applications},
  author={Schilit, Bill and Adams, Norman and Want, Roy},
  booktitle={1994 first workshop on mobile computing systems and applications},
  pages={85--90},
  year={1994},
  organization={IEEE}
}

@article{dey2001understanding,
  title={Understanding and using context},
  author={Dey, Anind K},
  journal={Personal and ubiquitous computing},
  volume={5},
  number={1},
  pages={4--7},
  year={2001},
  publisher={Springer}
}

@article{bulling2014tutorial,
author = {Bulling, Andreas and Blanke, Ulf and Schiele, Bernt},
title = {A tutorial on human activity recognition using body-worn inertial sensors},
year = {2014},
issue_date = {January 2014},
publisher = {Association for Computing Machinery},
address = {New York, NY, USA},
volume = {46},
number = {3},
issn = {0360-0300},
doi = {10.1145/2499621},
journal = {ACM Comput. Surv.},
month = jan,
}

@article{haresamudram2025past,
author = {Haresamudram, Harish and Tang, Chi Ian and Suh, Sungho and Lukowicz, Paul and Pl\"{o}tz, Thomas},
title = {Past, Present, and Future of Sensor-based Human Activity Recognition Using Wearables: A Surveying Tutorial on a Still Challenging Task},
year = {2025},
issue_date = {June 2025},
publisher = {Association for Computing Machinery},
address = {New York, NY, USA},
volume = {9},
number = {2},
doi = {10.1145/3729467},
journal = {Proc. ACM Interact. Mob. Wearable Ubiquitous Technol.},
month = jun,
articleno = {34},
numpages = {44}
}

@article{cai2025towards,
  title={Towards generalizable human activity recognition: A survey},
  author={Cai, Yize and Guo, Baoshen and Salim, Flora and Hong, Zhiqing},
  journal={arXiv preprint arXiv:2508.12213},
  year={2025}
}

@ARTICLE{kunze2014sensor,
  author={Kunze, Kai and Lukowicz, Paul},
  journal={IEEE Pervasive Computing}, 
  title={Sensor Placement Variations in Wearable Activity Recognition}, 
  year={2014},
  volume={13},
  number={4},
  pages={32-41},
  doi={10.1109/MPRV.2014.73}
  }

@article{chang2020systematic,
author = {Chang, Youngjae and Mathur, Akhil and Isopoussu, Anton and Song, Junehwa and Kawsar, Fahim},
title = {A Systematic Study of Unsupervised Domain Adaptation for Robust Human-Activity Recognition},
year = {2020},
issue_date = {March 2020},
publisher = {Association for Computing Machinery},
address = {New York, NY, USA},
volume = {4},
number = {1},
doi = {10.1145/3380985},
journal = {Proc. ACM Interact. Mob. Wearable Ubiquitous Technol.},
month = mar,
articleno = {39},
numpages = {30}
}

@inproceedings{stisen2015smart,
author = {Stisen, Allan and Blunck, Henrik and Bhattacharya, Sourav and Prentow, Thor Siiger and Kj\ae{}rgaard, Mikkel Baun and Dey, Anind and Sonne, Tobias and Jensen, Mads M\o{}ller},
title = {Smart Devices are Different: Assessing and MitigatingMobile Sensing Heterogeneities for Activity Recognition},
year = {2015},
isbn = {9781450336314},
publisher = {Association for Computing Machinery},
address = {New York, NY, USA},
doi = {10.1145/2809695.2809718},
booktitle = {Proceedings of the 13th ACM Conference on Embedded Networked Sensor Systems},
pages = {127–140},
numpages = {14},
location = {Seoul, South Korea},
series = {SenSys '15}
}

@article{bian2026foundation,
  title={Foundation Models Defining A New Era In Sensor-based Human Activity Recognition: A Survey And Outlook},
  author={Bian, Sizhen and Liu, Mengxi and Yuan, Siyu and Ray, Lala Shakti Swarup and Zhou, Bo and Guo, Bin and Yu, Zhiwen and Ploetz, Thomas and Lukowicz, Paul and Rey, Vitor Fortes},
  journal={arXiv preprint arXiv:2604.02711},
  year={2026}
}

@inproceedings{ma2024nymeria,
  title={Nymeria: A massive collection of multimodal egocentric daily motion in the wild},
  author={Ma, Lingni and Ye, Yuting and Hong, Fangzhou and Guzov, Vladimir and Jiang, Yifeng and Postyeni, Rowan and Pesqueira, Luis and Gamino, Alexander and Baiyya, Vijay and Kim, Hyo Jin and others},
  booktitle={European Conference on Computer Vision},
  pages={445--465},
  year={2024},
  organization={Springer}
}

@ARTICLE{oishi2025wimusim,
    
AUTHOR={Oishi, Nobuyuki  and Birch, Phil  and Roggen, Daniel  and Lago, Paula },
           
TITLE={WIMUSim: simulating realistic variabilities in wearable IMUs for human activity recognition},
          
JOURNAL={Frontiers in Computer Science},
          
VOLUME={Volume 7 - 2025},
  
YEAR={2025},
  
URL={https://www.frontiersin.org/journals/computer-science/articles/10.3389/fcomp.2025.1514933},
  
DOI={10.3389/fcomp.2025.1514933},
  
ISSN={2624-9898},
}

@article{zhang2024unimts,
  title={Unimts: Unified pre-training for motion time series},
  author={Zhang, Xiyuan and Teng, Diyan and Chowdhury, Ranak R and Li, Shuheng and Hong, Dezhi and Gupta, Rajesh K and Shang, Jingbo},
  journal={Advances in Neural Information Processing Systems},
  volume={37},
  pages={107469--107493},
  year={2024}
}

@article{leng2024imugpt2,
author = {Leng, Zikang and Bhattacharjee, Amitrajit and Rajasekhar, Hrudhai and Zhang, Lizhe and Bruda, Elizabeth and Kwon, Hyeokhyen and Pl\"{o}tz, Thomas},
title = {IMUGPT 2.0: Language-Based Cross Modality Transfer for Sensor-Based Human Activity Recognition},
year = {2024},
issue_date = {September 2024},
publisher = {Association for Computing Machinery},
address = {New York, NY, USA},
volume = {8},
number = {3},
doi = {10.1145/3678545},
journal = {Proc. ACM Interact. Mob. Wearable Ubiquitous Technol.},
month = sep,
articleno = {112},
numpages = {32}
}

@inproceedings{leng2023generating,
author = {Leng, Zikang and Kwon, Hyeokhyen and Ploetz, Thomas},
title = {Generating Virtual On-body Accelerometer Data from Virtual Textual Descriptions for Human Activity Recognition},
year = {2023},
isbn = {9798400701993},
publisher = {Association for Computing Machinery},
address = {New York, NY, USA},
doi = {10.1145/3594738.3611361},
booktitle = {Proceedings of the 2023 ACM International Symposium on Wearable Computers},
pages = {39–43},
numpages = {5},
location = {Cancun, Quintana Roo, Mexico},
series = {ISWC '23}
}

@article{wei2025one,
author = {Wei, Qingxin and Huang, Jiaming and Gao, Yi and Dong, Wei},
title = {One Model to Fit Them All: Universal IMU-based Human Activity Recognition with LLM-assisted Cross-dataset Representation},
year = {2025},
issue_date = {September 2025},
publisher = {Association for Computing Machinery},
address = {New York, NY, USA},
volume = {9},
number = {3},
doi = {10.1145/3749509},
journal = {Proc. ACM Interact. Mob. Wearable Ubiquitous Technol.},
month = sep,
articleno = {139},
numpages = {22}
}

@article{miao2026wonderwall,
author = {Miao, Shenghuan and Chen, Ling},
title = {Wonderwall: A Virtual-to-Real Foundation Model for IMU-based HAR},
year = {2026},
issue_date = {March 2026},
publisher = {Association for Computing Machinery},
address = {New York, NY, USA},
volume = {10},
number = {1},
doi = {10.1145/3789688},
journal = {Proc. ACM Interact. Mob. Wearable Ubiquitous Technol.},
month = mar,
articleno = {14},
numpages = {31}
}

@article{hong2024crosshar,
author = {Hong, Zhiqing and Li, Zelong and Zhong, Shuxin and Lyu, Wenjun and Wang, Haotian and Ding, Yi and He, Tian and Zhang, Desheng},
title = {CrossHAR: Generalizing Cross-dataset Human Activity Recognition via Hierarchical Self-Supervised Pretraining},
year = {2024},
issue_date = {June 2024},
publisher = {Association for Computing Machinery},
address = {New York, NY, USA},
volume = {8},
number = {2},
doi = {10.1145/3659597},
journal = {Proc. ACM Interact. Mob. Wearable Ubiquitous Technol.},
month = may,
articleno = {64},
numpages = {26}
}

@inproceedings{yan2023skeletonmae,
  title={Skeletonmae: graph-based masked autoencoder for skeleton sequence pre-training},
  author={Yan, Hong and Liu, Yang and Wei, Yushen and Li, Zhen and Li, Guanbin and Lin, Liang},
  booktitle={Proceedings of the IEEE/CVF international conference on computer vision},
  pages={5606--5618},
  year={2023}
}

@inproceedings{mao2023masked,
  title={Masked motion predictors are strong 3d action representation learners},
  author={Mao, Yunyao and Deng, Jiajun and Zhou, Wengang and Fang, Yao and Ouyang, Wanli and Li, Houqiang},
  booktitle={Proceedings of the IEEE/CVF International Conference on Computer Vision},
  pages={10181--10191},
  year={2023}
}

@inproceedings{huang2023graph,
title={Graph Contrastive Learning for Skeleton-based Action Recognition},
author={Xiaohu Huang and Hao Zhou and Jian Wang and Haocheng Feng and Junyu Han and Errui Ding and Jingdong Wang and Xinggang Wang and Wenyu Liu and Bin Feng},
booktitle={The Eleventh International Conference on Learning Representations },
year={2023},
url={https://openreview.net/forum?id=PLUXnnxUdr4}
}

@inproceedings{lin2023actionlet,
  title={Actionlet-dependent contrastive learning for unsupervised skeleton-based action recognition},
  author={Lin, Lilang and Zhang, Jiahang and Liu, Jiaying},
  booktitle={Proceedings of the IEEE/CVF Conference on Computer Vision and Pattern Recognition},
  pages={2363--2372},
  year={2023}
}

@inproceedings{hou2023graphmae2,
  title={Graphmae2: A decoding-enhanced masked self-supervised graph learner},
  author={Hou, Zhenyu and He, Yufei and Cen, Yukuo and Liu, Xiao and Dong, Yuxiao and Kharlamov, Evgeny and Tang, Jie},
  booktitle={Proceedings of the ACM web conference 2023},
  pages={737--746},
  year={2023}
}

@inproceedings{li20213d,
  title={3d human action representation learning via cross-view consistency pursuit},
  author={Li, Linguo and Wang, Minsi and Ni, Bingbing and Wang, Hang and Yang, Jiancheng and Zhang, Wenjun},
  booktitle={Proceedings of the IEEE/CVF conference on computer vision and pattern recognition},
  pages={4741--4750},
  year={2021}
}

@inproceedings{yan2018spatial,
  title={Spatial temporal graph convolutional networks for skeleton-based action recognition},
  author={Yan, Sijie and Xiong, Yuanjun and Lin, Dahua},
  booktitle={Proceedings of the AAAI conference on artificial intelligence},
  volume={32},
  number={1},
  year={2018}
}

@article{vaswani2017attention,
  title={Attention is all you need},
  author={Vaswani, Ashish and Shazeer, Noam and Parmar, Niki and Uszkoreit, Jakob and Jones, Llion and Gomez, Aidan N and Kaiser, {\L}ukasz and Polosukhin, Illia},
  journal={Advances in neural information processing systems},
  volume={30},
  year={2017}
}

@inproceedings{hong2025egolm,
  title={Egolm: Multi-modal language model of egocentric motions},
  author={Hong, Fangzhou and Guzov, Vladimir and Kim, Hyo Jin and Ye, Yuting and Newcombe, Richard and Liu, Ziwei and Ma, Lingni},
  booktitle={Proceedings of the Computer Vision and Pattern Recognition Conference},
  pages={5344--5354},
  year={2025}
}

@inproceedings{muennighoff2025generative,
title={Generative Representational Instruction Tuning},
author={Niklas Muennighoff and Hongjin SU and Liang Wang and Nan Yang and Furu Wei and Tao Yu and Amanpreet Singh and Douwe Kiela},
booktitle={The Thirteenth International Conference on Learning Representations},
year={2025},
url={https://openreview.net/forum?id=BC4lIvfSzv}
}

@inproceedings{lee2025nvembed,
title={{NV}-Embed: Improved Techniques for Training {LLM}s as Generalist Embedding Models},
author={Chankyu Lee and Rajarshi Roy and Mengyao Xu and Jonathan Raiman and Mohammad Shoeybi and Bryan Catanzaro and Wei Ping},
booktitle={The Thirteenth International Conference on Learning Representations},
year={2025},
url={https://openreview.net/forum?id=lgsyLSsDRe}
}

@inproceedings{yu2025cafe,
  title={Cafe: Unifying representation and generation with contrastive-autoregressive finetuning},
  author={Yu, Hao and Zhao, Zhuokai and Yan, Shen and Korycki, Lukasz and Wang, Jianyu and He, Baosheng and Liu, Jiayi and Zhang, Lizhu and Fan, Xiangjun and Yu, Hanchao},
  booktitle={Proceedings of the IEEE/CVF International Conference on Computer Vision},
  pages={6286--6297},
  year={2025}
}

@inproceedings{he2016deep,
  title={Deep residual learning for image recognition},
  author={He, Kaiming and Zhang, Xiangyu and Ren, Shaoqing and Sun, Jian},
  booktitle={Proceedings of the IEEE conference on computer vision and pattern recognition},
  pages={770--778},
  year={2016}
}

@inproceedings{tan2026bisecle,
title={Bisecle: Binding and Separation in Continual Learning for Video Language Understanding},
author={Yue Tan and Xiaoqian Hu and Hao Xue and Celso M de Melo and Flora D. Salim},
booktitle={The Thirty-ninth Annual Conference on Neural Information Processing Systems},
year={2025},
url={https://openreview.net/forum?id=o6keqobP13}
}

@article{liu2024gpt,
  title={GPT understands, too},
  author={Liu, Xiao and Zheng, Yanan and Du, Zhengxiao and Ding, Ming and Qian, Yujie and Yang, Zhilin and Tang, Jie},
  journal={AI open},
  volume={5},
  pages={208--215},
  year={2024},
  publisher={Elsevier}
}

@techreport{openai2025gpt54thinking,
  title        = {{GPT-5.4 Thinking System Card}},
  author       = {{OpenAI}},
  institution  = {{OpenAI}},
  year         = {2026},
  month        = mar,
  url          = {https://deploymentsafety.openai.com/gpt-5-4-thinking/gpt-5-4-thinking.pdf},
  note         = {System card}
}

@techreport{googledeepmind2026gemma4modelcard,
  title        = {{Gemma 4 Model Card}},
  author       = {{Google DeepMind}},
  year         = {2026},
  month        = apr,
  howpublished = {\url{https://ai.google.dev/gemma/docs/core/model_card_4}},
  note         = {Last updated: 2026-04-17}
}

@article{jiang2023motiongpt,
  title={Motiongpt: Human motion as a foreign language},
  author={Jiang, Biao and Chen, Xin and Liu, Wen and Yu, Jingyi and Yu, Gang and Chen, Tao},
  journal={Advances in Neural Information Processing Systems},
  volume={36},
  pages={20067--20079},
  year={2023}
}

@inproceedings{zhang2025sensorlm,
title={Sensor{LM}: Learning the Language of Wearable Sensors},
author={Yuwei Zhang and Kumar Ayush and Siyuan Qiao and A. Ali Heydari and Girish Narayanswamy and Maxwell A Xu and Ahmed Metwally and Jinhua Xu and Jake Garrison and Xuhai Xu and Tim Althoff and Yun Liu and Pushmeet Kohli and Jiening Zhan and Mark Malhotra and Shwetak Patel and Cecilia Mascolo and Xin Liu and Daniel McDuff and Yuzhe Yang},
booktitle={The Thirty-ninth Annual Conference on Neural Information Processing Systems},
year={2025},
url={https://openreview.net/forum?id=TrHeq0yFhv}
}

@inproceedings{roggen2010collecting,
  title={Collecting complex activity datasets in highly rich networked sensor environments},
  author={Roggen, Daniel and Calatroni, Alberto and Rossi, Mirco and Holleczek, Thomas and F{\"o}rster, Kilian and Tr{\"o}ster, Gerhard and Lukowicz, Paul and Bannach, David and Pirkl, Gerald and Ferscha, Alois and others},
  booktitle={2010 Seventh international conference on networked sensing systems (INSS)},
  pages={233--240},
  year={2010},
  organization={IEEE}
}

@inproceedings{anguita2013public,
  title={A public domain dataset for human activity recognition using smartphones.},
  author={Anguita, Davide and Ghio, Alessandro and Oneto, Luca and Parra, Xavier and Reyes-Ortiz, Jorge Luis and others},
  booktitle={Esann},
  volume={3},
  number={1},
  pages={3--4},
  year={2013}
}

@article{bhat2020w,
  title={w-HAR: An activity recognition dataset and framework using low-power wearable devices},
  author={Bhat, Ganapati and Tran, Nicholas and Shill, Holly and Ogras, Umit Y},
  journal={Sensors},
  volume={20},
  number={18},
  pages={5356},
  year={2020},
  publisher={MDPI}
}

@inproceedings{sztyler2016body,
  title={On-body localization of wearable devices: An investigation of position-aware activity recognition},
  author={Sztyler, Timo and Stuckenschmidt, Heiner},
  booktitle={2016 IEEE international conference on pervasive computing and communications (PerCom)},
  pages={1--9},
  year={2016},
  organization={IEEE}
}

@data{yan2021tnda,
doi = {10.21227/4epb-pg26},
url = {https://dx.doi.org/10.21227/4epb-pg26},
author = {Yan Yan and Dali Chen and Yushi Liu and Jinjin Zhao and Bo Wang and Xuankun Wu and Xiaohao Jiao and Yuqian Chen and Huihui Li and Xuchao Ren},
publisher = {IEEE Dataport},
title = {TNDA-HAR},
year = {2021} }

@inproceedings{grauman2024ego,
  title={Ego-exo4d: Understanding skilled human activity from first-and third-person perspectives},
  author={Grauman, Kristen and Westbury, Andrew and Torresani, Lorenzo and Kitani, Kris and Malik, Jitendra and Afouras, Triantafyllos and Ashutosh, Kumar and Baiyya, Vijay and Bansal, Siddhant and Boote, Bikram and others},
  booktitle={Proceedings of the IEEE/CVF Conference on Computer Vision and Pattern Recognition},
  pages={19383--19400},
  year={2024}
}

@inproceedings{yoshimura2024openpack,
  author={Yoshimura, Naoya and Morales, Jaime and Maekawa, Takuya and Hara, Takahiro},
  booktitle={2024 IEEE International Conference on Pervasive Computing and Communications (PerCom)}, 
  title={OpenPack: A Large-Scale Dataset for Recognizing Packaging Works in IoT-Enabled Logistic Environments}, 
  year={2024},
  volume={},
  number={},
  pages={90-97},
  doi={10.1109/PerCom59722.2024.10494448}}

@inproceedings{reiss2012introducing,
  title={Introducing a new benchmarked dataset for activity monitoring},
  author={Reiss, Attila and Stricker, Didier},
  booktitle={2012 16th international symposium on wearable computers},
  pages={108--109},
  year={2012},
  organization={IEEE}
}

@inproceedings{zhang2012usc,
  title={USC-HAD: A daily activity dataset for ubiquitous activity recognition using wearable sensors},
  author={Zhang, Mi and Sawchuk, Alexander A},
  booktitle={Proceedings of the 2012 ACM conference on ubiquitous computing},
  pages={1036--1043},
  year={2012}
}

@article{weiss2019wisdm,
  title={Wisdm smartphone and smartwatch activity and biometrics dataset},
  author={Weiss, Gary M},
  journal={UCI Machine Learning Repository: WISDM Smartphone and Smartwatch Activity and Biometrics Dataset Data Set},
  volume={7},
  number={133190-133202},
  pages={5},
  year={2019},
  publisher={Springer}
}

@article{altun2010comparative,
  title={Comparative study on classifying human activities with miniature inertial and magnetic sensors},
  author={Altun, Kerem and Barshan, Billur and Tun{\c{c}}el, Orkun},
  journal={Pattern Recognition},
  volume={43},
  number={10},
  pages={3605--3620},
  year={2010},
  publisher={Elsevier}
}

@inproceedings{chen2015utd,
  title={UTD-MHAD: A multimodal dataset for human action recognition utilizing a depth camera and a wearable inertial sensor},
  author={Chen, Chen and Jafari, Roozbeh and Kehtarnavaz, Nasser},
  booktitle={2015 IEEE International conference on image processing (ICIP)},
  pages={168--172},
  year={2015},
  organization={IEEE}
}

@inproceedings{grauman2022ego4d,
  title     = {{Ego4D}: Around the World in 3,000 Hours of Egocentric Video},
  author    = {Grauman, Kristen and Westbury, Andrew and Byrne, Eugene and Chavis, Zachary and Furnari, Antonino and Girdhar, Rohit and Hamburger, Jackson and Jiang, Hao and Liu, Miao and Liu, Xingyu and Martin, Miguel and others},
  booktitle = {Proceedings of the IEEE/CVF Conference on Computer Vision and Pattern Recognition},
  pages     = {18995--19012},
  year      = {2022}
}

@article{xu2024towards,
  author={Xu, Linfeng and Wu, Qingbo and Pan, Lili and Meng, Fanman and Li, Hongliang and He, Chiyuan and Wang, Hanxin and Cheng, Shaoxu and Dai, Yu},
  journal={IEEE Transactions on Multimedia}, 
  title={Towards Continual Egocentric Activity Recognition: A Multi-Modal Egocentric Activity Dataset for Continual Learning}, 
  year={2024},
  volume={26},
  number={},
  pages={2430-2443},
  doi={10.1109/TMM.2023.3295899}}

@inproceedings{girdhar2023imagebind,
  title     = {{ImageBind}: One Embedding Space to Bind Them All},
  author    = {Girdhar, Rohit and El-Nouby, Alaaeldin and Liu, Zhuang and Singh, Mannat and Alwala, Kalyan Vasudev and Joulin, Armand and Misra, Ishan},
  booktitle = {Proceedings of the IEEE/CVF Conference on Computer Vision and Pattern Recognition},
  pages     = {15180--15190},
  year      = {2023}
}

@inproceedings{moon2023imu2clip,
    title = "{IMU}2{CLIP}: Language-grounded Motion Sensor Translation with Multimodal Contrastive Learning",
    author = "Moon, Seungwhan  and
      Madotto, Andrea  and
      Lin, Zhaojiang  and
      Saraf, Aparajita  and
      Bearman, Amy  and
      Damavandi, Babak",
    editor = "Bouamor, Houda  and
      Pino, Juan  and
      Bali, Kalika",
    booktitle = "Findings of the Association for Computational Linguistics: EMNLP 2023",
    month = dec,
    year = "2023",
    address = "Singapore",
    publisher = "Association for Computational Linguistics",
    url = "https://aclanthology.org/2023.findings-emnlp.883/",
    doi = "10.18653/v1/2023.findings-emnlp.883",
    pages = "13246--13253",
}

@inproceedings{ji2024hargpt,
  author    = {Ji, Sijie and Zheng, Xinzhe and Wu, Chenshu},
  title     = {{HARGPT}: Are {LLM}s Zero-Shot Human Activity Recognizers?},
  booktitle = {2024 IEEE International Workshop on Foundation Models for Cyber-Physical Systems \& Internet of Things (FMSys)},
  year      = {2024},
  pages     = {38--43},
  doi       = {10.1109/FMSys62467.2024.00011}
}

@article{luo2026toward,
author = {Luo, Yunfei and Chen, Yuliang and Salekin, Asif and Rahman, Tauhidur},
title = {Toward Foundation Model for Multivariate Wearable Sensing of Physiological Signals},
year = {2026},
publisher = {Association for Computing Machinery},
address = {New York, NY, USA},
doi = {10.1145/3803808},
journal = {ACM Trans. Comput. Healthcare},
month = mar
}

@techreport{qwen2025qwen25technicalreport,
      title={Qwen2.5 Technical Report}, 
      author={Qwen and An Yang and Baosong Yang and Beichen Zhang and Binyuan Hui and Bo Zheng and Bowen Yu and Chengyuan Li and Dayiheng Liu and Fei Huang and Haoran Wei and Huan Lin and Jian Yang and Jianhong Tu and Jianwei Zhang and Jianxin Yang and Jiaxi Yang and Jingren Zhou and Junyang Lin and Kai Dang and Keming Lu and Keqin Bao and Kexin Yang and Le Yu and Mei Li and Mingfeng Xue and Pei Zhang and Qin Zhu and Rui Men and Runji Lin and Tianhao Li and Tianyi Tang and Tingyu Xia and Xingzhang Ren and Xuancheng Ren and Yang Fan and Yang Su and Yichang Zhang and Yu Wan and Yuqiong Liu and Zeyu Cui and Zhenru Zhang and Zihan Qiu},
      year={2025},
      eprint={2412.15115},
      archivePrefix={arXiv},
      primaryClass={cs.CL},
      url={https://arxiv.org/abs/2412.15115}, 
}

@article{mcinnes2018umap,
  title   = {{UMAP}: Uniform Manifold Approximation and Projection for Dimension Reduction},
  author  = {McInnes, Leland and Healy, John and Melville, James},
  journal = {arXiv preprint arXiv:1802.03426},
  year    = {2018},
  url     = {https://arxiv.org/abs/1802.03426}
}

@inproceedings{zhao2025swift,
  title     = {{SWIFT}: A Scalable Lightweight Infrastructure for Fine-Tuning},
  author    = {Zhao, Yuze and Huang, Jiaxing and Hu, Junjie and Wang, Xintao and Mao, Yuxuan and Zhang, Dingkang and Jiang, Zhenyu and Wu, Ziyi and Ai, Bin and Wang, Ang and Zhou, Wen},
  booktitle = {Proceedings of the AAAI Conference on Artificial Intelligence},
  volume    = {39},
  number    = {28},
  pages     = {29733--29735},
  year      = {2025}
}

@inproceedings{gao2021scaling,
    title = "Scaling Deep Contrastive Learning Batch Size under Memory Limited Setup",
    author = "Gao, Luyu  and
      Zhang, Yunyi  and
      Han, Jiawei  and
      Callan, Jamie",
    editor = "Rogers, Anna  and
      Calixto, Iacer  and
      Vuli{\'c}, Ivan  and
      Saphra, Naomi  and
      Kassner, Nora  and
      Camburu, Oana-Maria  and
      Bansal, Trapit  and
      Shwartz, Vered",
    booktitle = "Proceedings of the 6th Workshop on Representation Learning for NLP (RepL4NLP-2021)",
    month = aug,
    year = "2021",
    address = "Online",
    publisher = "Association for Computational Linguistics",
    doi = "10.18653/v1/2021.repl4nlp-1.31",
    pages = "316--321",
}

@misc{openai2025gptoss120bgptoss20b,
  title         = {{gpt-oss-120b \& gpt-oss-20b Model Card}},
  author        = {{OpenAI}},
  year          = {2025},
  eprint        = {2508.10925},
  archivePrefix = {arXiv},
  primaryClass  = {cs.CL},
  url           = {https://arxiv.org/abs/2508.10925}
}

@article{zhang2025qwen3textembedding,
  title         = {{Qwen3 Embedding}: Advancing Text Embedding and Reranking Through Foundation Models},
  author        = {Zhang, Yanzhao and Li, Mingxin and Long, Dingkun and Zhang, Xin and Lin, Huan and Yang, Baosong and Xie, Pengjun and Yang, An and Liu, Dayiheng and Lin, Junyang and Huang, Fei and Zhou, Jingren},
  journal       = {arXiv preprint arXiv:2506.05176},
  year          = {2025},
  eprint        = {2506.05176},
  archivePrefix = {arXiv},
  url           = {https://arxiv.org/abs/2506.05176}
}

@inproceedings{li2025sensorllm,
    title = "{S}ensor{LLM}: Aligning Large Language Models with Motion Sensors for Human Activity Recognition",
    author = "Li, Zechen  and
      Deldari, Shohreh  and
      Chen, Linyao  and
      Xue, Hao  and
      Salim, Flora D.",
    editor = "Christodoulopoulos, Christos  and
      Chakraborty, Tanmoy  and
      Rose, Carolyn  and
      Peng, Violet",
    booktitle = "Proceedings of the 2025 Conference on Empirical Methods in Natural Language Processing",
    month = nov,
    year = "2025",
    address = "Suzhou, China",
    publisher = "Association for Computational Linguistics",
    url = "https://aclanthology.org/2025.emnlp-main.19/",
    doi = "10.18653/v1/2025.emnlp-main.19",
    pages = "354--379",
    ISBN = "979-8-89176-332-6",
}

@inproceedings{
xu2025relcon,
title={RelCon: Relative Contrastive Learning for a Motion Foundation Model for Wearable Data},
author={Maxwell A Xu and Jaya Narain and Gregory Darnell and Haraldur T Hallgrimsson and Hyewon Jeong and Darren Forde and Richard Andres Fineman and Karthik Jayaraman Raghuram and James Matthew Rehg and Shirley You Ren},
booktitle={The Thirteenth International Conference on Learning Representations},
year={2025},
url={https://openreview.net/forum?id=k2uUeLCrQq}
}

@article{chen2025comodo,
  title={Comodo: Cross-modal video-to-imu distillation for efficient egocentric human activity recognition},
  author={Chen, Baiyu and Wongso, Wilson and Li, Zechen and Khaokaew, Yonchanok and Xue, Hao and Salim, Flora},
  journal={arXiv preprint arXiv:2503.07259},
  year={2025}
}

@inproceedings{
zhang2025mopformer,
title={Mo{PF}ormer: Motion-Primitive Transformer for Wearable-Sensor Activity Recognition},
author={Hao Zhang and Zhan Zhuang and Xuehao Wang and Xiaodong Yang and Yu Zhang},
booktitle={The Thirty-ninth Annual Conference on Neural Information Processing Systems},
year={2025},
url={https://openreview.net/forum?id=Ty9n72fZ1K}
}

@inproceedings{fortes2022learning,
author = {Fortes Rey, Vitor and Suh, Sungho and Lukowicz, Paul},
title = {Learning from the Best: Contrastive Representations Learning Across Sensor Locations for Wearable Activity Recognition},
year = {2022},
isbn = {9781450394246},
publisher = {Association for Computing Machinery},
address = {New York, NY, USA},
doi = {10.1145/3544794.3558464},
pages = {28–32},
numpages = {5},
location = {Cambridge, United Kingdom},
series = {ISWC '22}
}

@inproceedings{ray2025w2w,
author = {Ray, Lala Shakti Swarup and Zhou, Bo and Lukowicz, Paul},
title = {W2W: A Simulated Exploration of IMU Placement Across the Human Body for Designing Smarter Wearable},
year = {2025},
isbn = {9798400714818},
publisher = {Association for Computing Machinery},
address = {New York, NY, USA},
doi = {10.1145/3715071.3750417},
booktitle = {Proceedings of the 2025 ACM International Symposium on Wearable Computers},
pages = {170–176},
numpages = {7},
location = {Espoo, Finland},
series = {ISWC '25}
}

@inproceedings{zhou2025imucoco,
author = {Zhou, Haozhe and Arakawa, Riku and Agarwal, Yuvraj and Goel, Mayank},
title = {IMUCoCo: Enabling Flexible On-Body IMU Placement for Human Pose Estimation and Activity Recognition},
year = {2025},
isbn = {9798400720376},
publisher = {Association for Computing Machinery},
address = {New York, NY, USA},
doi = {10.1145/3746059.3747695},
booktitle = {Proceedings of the 38th Annual ACM Symposium on User Interface Software and Technology},
articleno = {91},
numpages = {16},
location = {
},
series = {UIST '25}
}

@inproceedings{asif2025llasa,
author = {Asif Imran Shouborno, Sheikh and Khan, Mohammad Nur Hossain and Biswas, Subrata and Islam, Bashima},
title = {LLaSA: A Sensor-Aware LLM for Natural Language Reasoning of Human Activity from IMU Data},
year = {2025},
isbn = {9798400714771},
publisher = {Association for Computing Machinery},
address = {New York, NY, USA},
doi = {10.1145/3714394.3756187},
booktitle = {Companion of the 2025 ACM International Joint Conference on Pervasive and Ubiquitous Computing},
pages = {893–899},
numpages = {7},
location = {Finland},
series = {UbiComp Companion '25}
}

@article{wang2024ubiphysio,
author = {Wang, Chongyang and Feng, Yuan and Zhong, Lingxiao and Zhu, Siyi and Zhang, Chi and Zheng, Siqi and Liang, Chen and Wang, Yuntao and He, Chengqi and Yu, Chun and Shi, Yuanchun},
title = {UbiPhysio: Support Daily Functioning, Fitness, and Rehabilitation with Action Understanding and Feedback in Natural Language},
year = {2024},
issue_date = {March 2024},
publisher = {Association for Computing Machinery},
address = {New York, NY, USA},
volume = {8},
number = {1},
doi = {10.1145/3643552},
journal = {Proc. ACM Interact. Mob. Wearable Ubiquitous Technol.},
month = mar,
articleno = {20},
numpages = {27}
}

@article{su2025imuzero,
author = {Su, Jie and Ge, Fengtong and Wen, Zhenyu and Li, Taotao and Bai, Yang and Zhou, Yejian and Zhang, Xiaoqin},
title = {IMUZero: Zero-Shot Human Activity Recognition by Language-Based Cross Modality Fusion},
year = {2025},
issue_date = {December 2025},
publisher = {Association for Computing Machinery},
address = {New York, NY, USA},
volume = {9},
number = {4},
doi = {10.1145/3770669},
journal = {Proc. ACM Interact. Mob. Wearable Ubiquitous Technol.},
month = dec,
articleno = {211},
numpages = {28}
}

@article{li2025zara,
  title   = {ZARA: Training-Free Motion Time-Series Reasoning via Evidence-Grounded LLM Agents},
  author  = {Li, Zechen and Chen, Baiyu and Xue, Hao and Salim, Flora D},
  journal = {arXiv preprint arXiv:2508.04038},
  year    = {2025}
}

@article{nguyen2026mobind,
  title   = {MoBind: Motion Binding for Fine-Grained IMU-Video Pose Alignment},
  author  = {Nguyen, Duc Duy and Chin, Tat-Jun and Hoai, Minh},
  journal = {arXiv preprint arXiv:2602.19004},
  year    = {2026}
}

@inproceedings{kinfu2025motionbind,
  title     = {MotionBind: Multi-Modal Human Motion Alignment for Retrieval, Recognition, and Generation},
  author    = {Kaleab A Kinfu and Rene Vidal},
  booktitle = {The Thirty-ninth Annual Conference on Neural Information Processing Systems},
  year      = {2025},
  url       = {https://openreview.net/forum?id=sUjwDdyspc}
}

@inproceedings{
zhou2026holollm,
title={Holo{LLM}: Multisensory Foundation Model for Language-Grounded Human Sensing and Reasoning},
author={Chuhao Zhou and Jianfei Yang},
booktitle={The Thirty-ninth Annual Conference on Neural Information Processing Systems},
year={2025},
url={https://openreview.net/forum?id=cHMP2IAhML}
}

@article{tong2022zero,
author = {Tong, Catherine and Ge, Jinchen and Lane, Nicholas D.},
title = {Zero-Shot Learning for IMU-Based Activity Recognition Using Video Embeddings},
year = {2022},
issue_date = {Dec 2021},
publisher = {Association for Computing Machinery},
address = {New York, NY, USA},
volume = {5},
number = {4},
doi = {10.1145/3494995},
journal = {Proc. ACM Interact. Mob. Wearable Ubiquitous Technol.},
month = dec,
articleno = {180},
numpages = {23}
}

@inproceedings{
he2021deberta,
title={{\{}DEBERTA{\}}: {\{}DECODING{\}}-{\{}ENHANCED{\}} {\{}BERT{\}} {\{}WITH{\}} {\{}DISENTANGLED{\}} {\{}ATTENTION{\}}},
author={Pengcheng He and Xiaodong Liu and Jianfeng Gao and Weizhu Chen},
booktitle={International Conference on Learning Representations},
year={2021},
url={https://openreview.net/forum?id=XPZIaotutsD}
}

@Article{ordonez2016deep,
AUTHOR = {Ordóñez, Francisco Javier and Roggen, Daniel},
TITLE = {Deep Convolutional and LSTM Recurrent Neural Networks for Multimodal Wearable Activity Recognition},
JOURNAL = {Sensors},
VOLUME = {16},
YEAR = {2016},
NUMBER = {1},
ARTICLE-NUMBER = {115},
URL = {https://www.mdpi.com/1424-8220/16/1/115},
PubMedID = {26797612},
ISSN = {1424-8220},
DOI = {10.3390/s16010115}
}

@article{feofanov2026mantisv2,
  title={MantisV2: Closing the Zero-Shot Gap in Time Series Classification with Synthetic Data and Test-Time Strategies},
  author={Feofanov, Vasilii and Wen, Songkang and Zhang, Jianfeng and Pan, Lujia and Redko, Ievgen},
  journal={arXiv preprint arXiv:2602.17868},
  year={2026}
}

@inproceedings{wang2025ego4o,
  title={Ego4o: Egocentric human motion capture and understanding from multi-modal input},
  author={Wang, Jian and Dabral, Rishabh and Luvizon, Diogo and Cao, Zhe and Liu, Lingjie and Beeler, Thabo and Theobalt, Christian},
  booktitle={Proceedings of the Computer Vision and Pattern Recognition Conference},
  pages={22668--22679},
  year={2025}
}

@article{straczkiewicz2021systematic,
  author  = {Straczkiewicz, Marcin and James, Peter and Onnela, Jukka-Pekka},
  title   = {A systematic review of smartphone-based human activity recognition methods for health research},
  journal = {npj Digital Medicine},
  year    = {2021},
  volume  = {4},
  number  = {1},
  pages   = {148},
  doi     = {10.1038/s41746-021-00514-4},
  url     = {https://doi.org/10.1038/s41746-021-00514-4},
  issn    = {2398-6352}
}

@article{li2026glucofm,
  title={GlucoFM: A Dual-Stream Foundation Model for Continuous Glucose Monitoring},
  author={Li, Zechen and Natarajan, Keerthana and Zhang, Weizhi and Zhou, Menglian and Lee, Simon A and Zhang, Yuwei and Xu, Maxwell A and Esmaeilpour, Zeinab and Salim, Flora D and Malhotra, Mark and others},
  journal={arXiv preprint arXiv:2605.30865},
  year={2026}
}

@article{li2026audiofm,
  title={AudioFM: Towards General Purpose Audio Representation Learning},
  author={Li, Yamin and Zhang, Ruidong and Wang, Anran and Fonseca, Eduardo and Ellis, Dan and Li, Zechen and Lee, Simon A and Pang, Tao and Garrison, Jake and Xu, Maxwell A and others},
  journal={arXiv preprint arXiv:2610.34183},
  year={2026}
}

@dataset{palermo2022cipdatabase,
    author       = {Manuel Palermo and Sara Cerqueira and João André and António Pereira and Cristina P. Santos},
    title        = {Complete Inertial Pose (CIP) Dataset},
    year         = {2022},
    publisher    = {Zenodo},
    doi          = {10.5281/zenodo.5801928},
    url          = {https://doi.org/10.5281/zenodo.5801928}
}
